\documentclass[journal]{IEEEtran}
\usepackage[acronym, shortcuts]{glossaries}
\usepackage{graphicx}
\usepackage{pifont}
\usepackage{gensymb}
\usepackage{bbding}
\usepackage{subcaption}
\usepackage{booktabs}
\usepackage{hyperref} 
\usepackage[font=footnotesize]{caption}    % Applies to all captions
\newacronym{lr}{LR}{Low Resolution}
\newacronym{hr}{HR}{High Resolution}
\newacronym{nn}{NN}{Neural Network}
\newacronym{sr}{SR}{Super-Resolution}
\newacronym{sisr}{SISR}{Single-Image Super-Resolution}
\newacronym{rs}{RS}{Remote Sensing}
\newacronym{dl}{DL}{Deep Learning}
\newacronym{da}{DA}{Domain Adaptation}
\newacronym{td}{TD}{top-down}
\newacronym{bu}{BU}{bottom-up}
\newacronym{mae}{MAE}{Mean Absolute Error}
\newacronym{mse}{MSE}{Mean Squared Error}
\newacronym{ssim}{SSIM}{Structural Similarity Index Measure}
\newacronym{nmse}{NMSE}{Normalized Mean Squared Error}
\newacronym{uiqi}{UIQI}{Universal Image Quality Index}
\newacronym{scc}{SCC}{Spatial Correlation Coefficient}
\newacronym{maxae}{MaxAE}{Maximum Absolute Error}
\newacronym{megan}{MEGAN}{Model of Emissions of Gases and Aerosols from Nature}
\newacronym{form}{HCHO}{formaldehyde}
\newacronym{bvoc}{BVOC}{Biogenic Volatile Organic Compound}
\newacronym{surfex}{SURFEX}{Surface Externalisée}
\newacronym{ctm}{CTM}{Chemical Transport Model}
\newacronym{ml}{ML}{Machine Learning}
\newacronym{tropomi}{TROPOMI}{TROPOspheric Monitoring Instrument}
\newacronym{era5}{ERA5}{ECMWF Reanalysis v5}
\newacronym{misr}{MISR}{Multi-Image Super-Resolution}
\newacronym{medfld}{\textit{Med}}{Mediterranean climate cluster}
\newacronym{san}{SAN}{Second-order Attention Network}
\newacronym{hat}{HAT}{Hybrid Attention Transformer}
\newacronym{srcnn}{SRCNN}{SR Convolutional Neural Network}
\newacronym{lai}{LAI}{Leaf Area Index}
\newacronym{esm}{ESM}{Earth System Model}
\newacronym{pcc}{PCC}{Pearson Correlation Coefficient}
\newacronym{cnn}{CNN}{Convolutional Neural Network}
\newacronym{seeds}{SEEDS}{Sentinel EO-based Emission and Deposition Service}
\newacronym{magritte}{MAGRITTE}{Model of Atmospheric composition at Global and Regional scales using Inversion Techniques for Trace gas Emissions}
\newacronym{nir}{NIR}{Normalized Improvement Ratio}
\newacronym{eo}{EO}{Earth Observation}
\newacronym{cl}{CL}{Cropland}
\newacronym{tc}{TC}{Tree Cover}
\newacronym{lc}{LC}{Land Cover}
\newacronym{ocab}{OCAB}{Overlapping Cross-Attention Block}
\newacronym{rlnl}{RL-NL}{Region-level Non-Local}
\newacronym{ce}{CE}{Conditional Entropy}
\newacronym{psnr}{PSNR}{Peak Signal-to-Noise Ratio}
\newacronym{esa}{ESA}{European Space Agency}

\newcommand{\HR}{\mathrm{\mathbf{I}_{HR}}}
\newcommand{\HRhat}{\mathrm{\mathbf{\hat{I}}_{HR}}}
\newcommand{\LR}{\mathrm{\mathbf{I}_{LR}}}

\newcommand{\DLR}{\mathcal{D}_{\mathrm{LR}}}

\newcommand{\HRt}{\mathrm{\mathbf{T}_{HR}}}
\newcommand{\HRhatt}{\mathrm{\mathbf{\hat{T}}_{HR}}}
\newcommand{\LRt}{\mathrm{\mathbf{T}_{LR}}}
\newcommand{\HRhatRef}{\mathrm{\mathbf{\hat{T}}_{HR}^{Ref.}}}

\newcommand{\Confone}{\{\text{Isop.}\}}
\newcommand{\Conftwocl}{\{\text{Isop., CL}\}}
\newcommand{\Conftwotc}{\{\text{Isop., TC}\}}
\newcommand{\Confthree}{\{\text{Isop., CL, TC}\}}

\newcommand{\T}{\mathcal{T}}
\newcommand{\Tinv}{\mathcal{T}^{-1}}
\newcommand{\Net}{\mathcal{N}}

\newcommand{\AllSet}{\mathcal{S}}
\newcommand{\Foldn}{\mathcal{S}_{\setminus cc}}

\newcommand{\Foldone}{\mathcal{S}_{\setminus Cfb}}
\newcommand{\Foldtwo}{\mathcal{S}_{\setminus Dfb}}
\newcommand{\Foldthree}{\mathcal{S}_{\setminus Dfc}}
\newcommand{\Foldfour}{\mathcal{S}_{\setminus Med}}

\newcommand{\Climone}{Cfb}
\newcommand{\Climtwo}{Dfb}
\newcommand{\Climthree}{Dfc}
\newcommand{\Climfour}{Med}

\newcommand{\seeds}{\text{TD-TROPO-010}}
\newcommand{\esawc}{\text{LC-ESA}}
\newcommand{\kgclim}{\text{CZ-KG-010}}
\newcommand{\laiseeds}{\text{LAI-TROPO-010}}

\newcommand{\mse}{\operatorname{MSE}}

\newcommand{\Entr}{\mathrm{H}}

\begin{document}
\title{Leveraging Land Cover Priors for Isoprene Emission Super-Resolution}

\author{Christopher Ummerle, Antonio Giganti, Sara Mandelli, Paolo Bestagini, and Stefano Tubaro% <-this % stops a space
\thanks{All authors are from the Department of Electronics, Information and Bioengineering - Politecnico di Milano - Milan, Italy\\
Image and Sound Processing Lab (ISPL) - Politecnico di Milano - Milan, Italy}% <-this % stops a space
\thanks{This work was supported by the Italian Ministry of University and Research (MUR) and the European Union (EU) under the PON/REACT project.}% <-this % stops a space
\thanks{The implementation code of the presented methodology is available at \url{https://github.com/polimi-ispl/sr-bvoc-lc}. The $\seeds$ isoprene emission inventory is available at \url{https://www.seedsproject.eu/data/top-down-isoprene-emissions}. The $\laiseeds$ inventory is available at \url{https://www.seedsproject.eu/data/lai-ol}. The $\esawc$ inventory is available at \url{https://zenodo.org/records/7254221}. The $\kgclim$ inventory is available at \url{https://www.gloh2o.org/koppen/}. No restrictions apply to the availability of these data.}}


%\markboth{Journal of \LaTeX\ Class Files,~Vol.~XX, No.~X, Month~XXX}%
%{Ummerle \MakeLowercase{\textit{et al.}}: Leveraging Land Cover Priors for Isoprene Emission Super-Resolution}

\maketitle

\begin{abstract}
Remote sensing plays a crucial role in monitoring Earth's ecosystems, yet satellite-derived data often suffer from limited spatial resolution, restricting their applicability in atmospheric modeling and climate research. In this work, we propose a deep learning-based Super-Resolution (SR) framework that leverages land cover information to enhance the spatial accuracy of Biogenic Volatile Organic Compounds (BVOCs) emissions, with a particular focus on isoprene. Our approach integrates land cover priors as emission drivers, capturing spatial patterns more effectively than traditional methods.
We evaluate the model's performance across various climate conditions and analyze statistical correlations between isoprene emissions and key environmental information such as cropland and tree cover data. Additionally, we assess the generalization capabilities of our SR model by applying it to unseen climate zones and geographical regions. Experimental results demonstrate that incorporating land cover data significantly improves emission SR accuracy, particularly in heterogeneous landscapes.
This study contributes to atmospheric chemistry and climate modeling by providing a cost-effective, data-driven approach to refining BVOC emission maps. The proposed method enhances the usability of satellite-based emissions data, supporting applications in air quality forecasting, climate impact assessments, and environmental studies.
\end{abstract}


\section{Introduction}
\label{sec:introduction}
%%%% CONTEXT %%%%
\gls{rs} technology is crucial for monitoring and advancing our understanding of Earth Systems. Particularly, our understanding of the biotic compartments of ecosystems, integral to our living environment and agriculture, faces increasing uncertainties from global warming-induced climate change. To effectively mitigate and adapt to future impacts and threats, the availability of accurate, precise, and timely data on environmental changes is essential.

Satellites offer an effective technology for \gls{eo}, providing long-term, precise, and global measurements without extensive logistics and operational support. Once deployed, they continuously supply valuable data over vast areas, supporting ecological and agricultural monitoring~\cite{chen_book_rs_2024, tuia_ai4eo_2024}. However, satellite observations sometimes fall short of advanced application requirements regarding spatial and temporal resolution, limiting their applicability for specific uses, like wildfire management, air quality monitoring, or precision agriculture~\cite{reichstein_dl_nature_2019, sdraka_dl_sr_rs_2022, siddique_no2_satellite_review_2024}. 

%%%% SR %%%%
A practical approach to enhance satellite data without launching new satellite missions for deploying new sensors is using \gls{sr} techniques.
Over the past few decades, the use of \gls{dl}-based methods for \gls{sr} techniques has been steadily increasing, proving to be more effective than traditional statistical approaches in capturing local-scale patterns, offering a cost-efficient way to improve spatial resolution~\cite{hobeichi_ml_dyn_downscaling_2024, rampal_ai4downscale_climate_2024, sokhi_aq_research_2022}.

These advancements have significantly contributed to various climate-related studies~\cite{sdraka_dl_sr_rs_2022, reichstein_dl_nature_2019, park_downscaling_earth_dl_2022}. 
The high cost and societal importance of atmospheric modeling and observation systems further highlight the promise of \gls{dl}-based \gls{sr} techniques. In particular, climate data are challenging to super-resolve due to their sparsity and often skewed distributions, which \gls{dl}-based \gls{sr} techniques can handle better than traditional approaches~\cite{medina_cordex_multi_downscale_2022, hobeichi_ml_dyn_downscaling_2024}. Additionally, \gls{sr} models exhibit strong adaptability, enabling their application across various climates, temporal scales, and geographic locations when adequately trained~\cite{medina_xai_dl_ctm_2024, chiang_climate_dowscale_dl_2024, wang_da_aereosol_2022, yadav_deepaq_2023}.
Indeed, \gls{dl}-based \gls{sr} methods have already demonstrated significant success in Earth sciences, improving the spatial and temporal resolution of climate data derived from satellite observations and chemical transport model simulations~\cite{geiss_chem_sim_sr_2022, sdraka_dl_sr_rs_2022, china_sisr_overview_2021, reichstein_dl_nature_2019, tuia_ai4eo_2024}, driving progress in multiple research domains~\cite{siddique_no2_satellite_review_2024, sokhi_aq_research_2022, medina_cordex_multi_downscale_2022, eyring_ai4climate_2024, materia_ai_prediction_extremes_2023}. 

%%%% BVOC %%%%
In the areas of air quality and climate science, \glspl{bvoc} play a crucial role in processes impacting atmospheric composition, air quality, cloud formation, and climate regulation through their interactions within the biosphere and atmosphere, ultimately influencing both local ecosystems and global climate processes~\cite{dp2024isopreneCMIP6, weber_chemistry-driven_2022, akira_exchanges_bvoc_2021, laothawornkitkul_bvoc_2009, mircea_vegetation_urban_2023}. 
Among the \glspl{bvoc}, isoprene is by far the most important, accounting for nearly half of annual \gls{bvoc} emissions, thus playing a prominent role in atmospheric chemistry~\cite{bauwens_past_future_bvoc_mohycan_2018, ashworth_book_bvoc_2013, cai_bvoc_scientometric_2021, guenther_bvoc_1995}.

\gls{dl}-based \gls{sr} holds significant promise for enhancing variables derived from \gls{eo} data, as it manages non-linearity and complex patterns better than traditional models~\cite{chiang_climate_dowscale_dl_2024, lai_ai4climate_2024, rampal_ai4downscale_climate_2024}. A richer diversity of information further strengthens these \gls{sr} techniques \cite{sdraka_dl_sr_rs_2022, eyring_ai4climate_2024}. Since \gls{eo} data are available in many modalities from a wide variety of \gls{rs} missions and often represent individually valuable information, their fusion can offer even greater potential~\cite{giganti2023eusipco, vandal_climate_sr_2017, lloyd_misr_optically_2022, wang_precipitation_sr_2021, wu_mlp_fusion_2025, sdraka_dl_sr_rs_2022}. This integration is particularly interesting for interconnected biotic processes that influence diverse aspects of the environment. 

%%%% OUR WORK %%%%
In this work, we explore the integration of \gls{lc} information as a high-level representation of vegetation types into \gls{sr} models for biogenic emissions. This approach aims to enhance the spatial accuracy and applicability of \gls{bvoc}-related insights.
Among \glspl{bvoc}, isoprene is the most significant in terms of global emissions and atmospheric impact~\cite{ashworth_book_bvoc_2013, guenther_megan3_2020, silibello_bvoc_south_italy_2022}. 
Given that isoprene is our primary focus, we incorporate \gls{lc} priors into the network, analyze its performance under different climate conditions, and conduct statistical analyses of isoprene emissions, considering potential drivers such as \gls{cl} and \gls{tc} data. Additionally, we investigate the generalization capabilities of our method, particularly in super-resolving patches from unseen climate zones and geographical regions, where performance is more limited.

Our work introduces a novel approach for super-resolving isoprene emissions by integrating \gls{lc} data as a key prior. The primary contributions of this study include:

\begin{itemize}
\item Land Cover-Informed \gls{sr}: We propose a \gls{dl}-based \gls{sr} framework that incorporates \gls{lc} information to enhance the spatial pattern of \gls{bvoc} emissions, particularly isoprene.  
\item High-Resolution Satellite-Derived Inventory: We leverage an extremely recent high-resolution satellite-derived isoprene emission inventory to ensure accurate and up-to-date emission data for model training and evaluation.  
\item Recent Super-Resolution Network: We employ \gls{hat} a recent transformer-based \gls{sr} network, to super-resolve isoprene emissions with improved accuracy and spatial detail.  
\item Climate-Aware Performance Evaluation: We evaluate the model's performance across distinct climate classes by training the system on climate-specific conditions, analyzing robustness across diverse environmental settings.
\item Generalization Across Unseen Regions: We investigate the model's ability to generalize by super-resolving isoprene emissions in unseen climate conditions and geographical regions, identifying key challenges and potential improvements for robust performance.  
\end{itemize}

Our method represents a significant advancement for the atmospheric chemistry and climate modeling communities by improving the spatial accuracy of isoprene emission estimates, thereby enhancing air quality modeling and climate simulations.

The remainder of this paper is structured as follows: Section~\ref{sec:related_works} reviews related work in \gls{sr} for climate data. Section~\ref{sec:materials} describes the inventories used for validation. Section~\ref{sec:methods} details our proposed methodology. Section~\ref{sec:experimental_dataset} describes the data preprocessing and the experimental dataset construction. Section~\ref{sec:experimental_setup} report details regarding the experimental settings and 
Section~\ref{sec:results} contains an extensive experimental evaluation for validating our approach, also investigating key challenges related to generalization in emission \gls{sr}. Finally, Section~\ref{sec:discussion} and Section~\ref{sec:conclusions} conclude the work and discuss possible future directions.
\section{Related Works}
\label{sec:related_works}
Most existing \gls{sr} algorithms utilize proxy data and classical statistical techniques to enhance the resolution of satellite observations~\cite{ramacher_urbem_2021, marongiu_rf_air_ita_2024, mu_downscale_aq_2022, nuterman_downscale_atmo_2021, liu_air_pollution_2019}. Some integrate \gls{ml} and \gls{dl}~\cite{quesada_cordex_2023, li_improving_aq_2023}, while only a few focus specifically on emissions data~\cite{rampal_ai4downscale_climate_2024, hobeichi_ml_dyn_downscaling_2024}.

\gls{cnn}-based \gls{sr} approaches have been successfully applied across various climate science domains, including wind and solar modeling, satellite \gls{rs}, and atmospheric chemistry simulations~\cite{brecht_sisr_wind_2023, geiss_chem_sim_sr_2022, passarella_esm_fsrcnn_2022}. The \gls{sr} of land and sea surface temperatures, as well as precipitation, has also been extensively studied~\cite{medina_dl_stat_downscaling_2020, lloyd_misr_optically_2022, yingkai_dl_downscale_precipitation_2020, brecht_sisr_wind_2023, geiss_chem_sim_sr_2022, nguyen_sisr_lst_2022}. 

Most existing research primarily utilizes data from a single observation~\cite{brecht_sisr_wind_2023, nguyen_sisr_lst_2022, yasuda_micrometeorology_2022, stengel_solar_wind_sr_2020, giganti2023icip, brecht_sisr_wind_2023}, framing the problem as a \gls{sisr} task. However, integrating multiple observations from the same or related domains can enhance \gls{sr} performance~\cite{tr_misr_2022, misr_3drrdb_2022, aiello2024pretr-sr}, shifting the problem to a \gls{misr} setting~\cite{miller_rs_ts_2024, salvetti_misr_2020, razzak_misr_building_2023}.

For example, authors in~\cite{lloyd_misr_optically_2022} improve sea surface temperature resolution by combining optical and thermal imagery. 
The correlations between optical and thermal imagery from Sentinel-3 for sea and land surface \gls{sr} temperature is also explored in~\cite{ping_sst_2021} and~\cite{izumi_sisr_sst_2022}. Authors in~\cite{tian_salinity_2022} reconstruct more detailed ocean subsurface salinity by integrating data on sea surface temperature, sea surface wind, and other \gls{rs} information. Likewise, in~\cite{liu_air_pollution_2019}, the authors leverage external factors and spatial-temporal relationships to enhance air pollution data resolution. In contrast, the work in~\cite{wang_precipitation_sr_2021} demonstrates that combining \gls{hr} topography maps with \gls{lr} precipitation, sea level pressure, and air temperature maps improves \gls{sr} performance.

Recent research efforts are also focused on super-resolving multiple climate variables using an ensemble of maps from different \gls{esm} simulations~\cite{damiani_exploring_2024, oyama_deep_2023, kumar_modern_2023}. \cite{mardani_diff_model_atmo_down_2024} adopted a generative diffusion model~\cite{ho_diffusion_2020} to super-resolve several climate variables from the \gls{era5} dataset. Furthermore, authors is~\cite{vandal_climate_sr_2017} propose a \gls{sr} model for \gls{esm} data, using a \gls{srcnn} architecture to enhance precipitation field resolution with additional elevation data. Similarly, in~\cite{passarella_esm_fsrcnn_2022}, authors develop a network to super-resolve multiple climate variables by combining them to reconstruct \gls{hr} monthly averaged climate maps.
Some approaches further enforce adherence to known physical laws within the \gls{sr} algorithm for climate variables~\cite{geiss_chem_sim_sr_2022, oyama_dl_climate_2023}.

% For instance, \cite{manzhu_dl_no2_2021} increased the resolution of NO\textsubscript{2} emission maps from tropospheric monitoring instruments by integrating \gls{lr} target data with in-situ observations and incorporating geographical and climate variables. Similarly, \cite{li_improving_aq_2023} enhanced the spatial resolution of multiple air pollutants by applying graph convolution to model the spatiotemporal dynamics of pollutants, encoding their local spreading characteristics. 

% \subsection{Super-Resolution of BVOC Emission Maps}
Recently, \gls{dl}-based \gls{sr} methods were also proposed to super-resolve \gls{bvoc} emission. Authors in~\cite{giganti2023icip} propose a framework to enhance isoprene emission spatially. In addition, they found that aggregating the emission data from multiple, especially uncorrelated, \gls{bvoc} compounds significantly enhances the \gls{sr} process by preserving spatial patterns
and improving fine-scale structures~\cite{giganti2023eusipco}. Authors in~\cite{giganti2023igarss} address the \gls{sr} of isoprene emission maps derived from satellite observations through the integration of \gls{sr} and \gls{da} techniques in a data scarcity scenario. Lastly, in~\cite{giganti2024mdpi}, they propose to exploit the more detailed isoprene emission from numerical models to estimate better a fine-grained version of isoprene emission coming from satellite observation proposing an \gls{da} framework based on CycleGAN~\cite{zhu_cycle-gan_2017}. In addition, they propose multiple losses in the training pipeline to reduce the distribution shift among the simulated and satellite-derived emissions, extending model applicability to real-world satellite-derived \gls{bvoc} maps.

\section{Materials}
\label{sec:materials}
% In this work, we use isoprene emissions data and land cover information to enhance the emissions \gls{sr} process with semantic relations. Therefore, we aim to employ data that most accurately mirrors actual conditions and better contextualizes isoprene emission patterns.

\subsection{Isoprene Inventory}
As isoprene emissions, we adopt the most up-to-date satellite-derived, i.e., \gls{td} approach, inventory available in the literature~\cite{isoprene_seeds_2024}, distributed among the \gls{seeds} project, a European Union-funded initiative conducted from January 1, 2021, to December 31, 2023~\cite{seeds_achievements_2024}. \gls{seeds} offers a wide range of emissions and deposition products, covering pollutants such as NOx, ammonia, biomass burning emissions, ozone, and \glspl{bvoc}. Of particular relevance to this research is the isoprene emissions inventory, which provides emissions with a very high spatial resolution of up to $0.1\degree \times 0.1\degree$, approximately $10\text{km} \times 10\text{km}$. Isoprene emissions are derived from an inversion with the \gls{magritte} v1.1 \gls{ctm}~\cite{muller_magritte_2019} and its adjoint. They are constrained by \gls{form} vertical columns data retrieved using the \gls{tropomi} instrument on board the Sentinel-5P satellite. 
The assimilated isoprene emissions cover the European area from 
2018 to 2022, with daily average profiles.
We refer to this inventory as $\seeds$.

\subsection{LAI Inventory}
\label{ssec:lai}
Vegetation parameters are also made available through the \gls{seeds} project to support understanding of factors that could influence biogenic emissions. These inventories include the \gls{lai}~\cite{lai_seeds_2024}. \gls{lai} reports the total leaf area per ground unit area, providing insights into vegetation density and seasonal dynamics. 
\gls{lai} information are computed using the \gls{surfex} land surface model~\cite{surfex_model_2013}.
%The model SEEDS uses simulates \gls{lai} dynamically based on observed vegetation types, sunlight exposure, and weather conditions. 
This offers beneficial information for isoprene release processes as \gls{lai} directly correlates with the presence of leaves, where most \glspl{bvoc}, like isoprene, are emitted~\cite{mircea_vegetation_urban_2023, silibello_bvoc_south_italy_2022}.
As for the isoprene emission, \gls{lai} maps cover Europe from 2018 to 2022, with a $0.1\degree \times 0.1\degree$ spatial resolution. The information is daily averaged.
We refer to this inventory as $\laiseeds$.

\subsection{Land Cover Data}
\label{ssec:land_cover}
\glsreset{cl}
\glsreset{tc}
Since isoprene emissions exhibit a high spatial heterogeneity, \gls{lc} data from the \gls{esa} WorldCover 2021 v200~\cite{esa_worldcover_v2_2021} are used. This inventory provides \gls{lc} class maps at very high spatial resolution, including 11 different classes of vegetation types and non-vegetative species, based on Sentinel-1 and 2 data.
%such as tree cover, shrubland, grassland, and cropland, 
%like built-up areas, water bodies, snow, and ice. 
The classifications in this dataset provide spatial information on \gls{lc} types relevant to \gls{bvoc} emissions, i.e., \gls{cl} and \gls{tc}. These classifications can identify the spatial distribution of vegetation types linked to \gls{bvoc} release and could contribute information about emission patterns dependent on \gls{lc}~\cite{ciccioli_bvoc_2023, guenther_megan3_2020}. The \gls{esa} WorldCover dataset only captures a single temporal state (released 2021, v200), but since \gls{lc} changes occur relatively slowly~\cite{opacka_isoprene_2021}, it is assumed to be valid over the full observation period of the $\seeds$~\cite{esa_worldcover_v2_2021} isoprene inventory.
\gls{esa} WorldCover maps cover the entire Earth's surface, with a $0.0001\degree \times 0.0001\degree$ spatial resolution, referred to as 2021.
We refer to this inventory as $\esawc$.

\subsection{Climate Class Data}
\label{ssec:climate_zones}
To capture the isoprene spatial variability driven by environmental factors 
%while enabling region-specific insights into model performance
, we adopt a recent climate zone dataset presented in~\cite{beck_kg_climate_2023} based on the Köppen-Geiger climate classification, a widely used system for categorizing global land climates. It divides climates into five major classes and 30 subclasses based on thresholds and seasonal patterns of monthly air temperature and precipitation. 

Since it has been observed that regions within the same climate class tend to share similar vegetation characteristics~\cite{beck_kg_climate_2023, opacka_isoprene_2021}, this information can be used to separate the study area into sub-areas which are expected to have similar species and thus similar isoprene emission mechanisms. 
The available climate maps cover the entire Earth's surface, with a maximum of $0.01\degree \times 0.01\degree$ spatial resolution and include six approximately 30-year periods from 1901 to 2099, each represented as a single map. 
We refer to this inventory as $\kgclim$.

%In Figure~\ref{fig:datasets}, we report example maps extracted from the different datasets and inventories adopted in this study. In addition, 

% Table~\ref{tab:datasets} reports additional details of this dataset and inventories. 

% \begin{table}[]
%     \centering
%     \begin{tabular}{c|c}
%          &  \\
%          & 
%     \end{tabular}
%     \caption{DATA SPECS of the different datasets and inventories adopted in this study. As our previous MDPI.}
%     \label{tab:datasets}
% \end{table}

\section{Methods}
\label{sec:methods}
% This section presents the proposed methodology for the super-resolving isoprene emission maps. We first introduce the proposed model architectures, then the experimental dataset's generation process and the hold-out sets for generalization studies. Next, we discuss the training settings and conclude with the evaluation metrics used to assess model performance.

\subsection{Leveraging Land Cover Priors for Isoprene Emission Super-Resolution}
\label{ssec:proposed_method}

% OLD VERSION-----------------------------------------------
% In this work, we propose to super-resolve \gls{lr} isoprene emission maps by leveraging the inherent relationships between isoprene and the semantic information extracted from land cover data.
% These data represent an important class of drivers for biogenic emissions like isoprene, as they capture key environmental and ecological factors influencing emission patterns. Leveraging the inherent relationships between land cover characteristics and emissions can improve modeling accuracy, offering more profound insights into the processes that drive isoprene emissions~\cite{penuelas_bvocs_2010, stavrakou_isoprene_asia_2014, ciccioli_bvoc_2023, ashworth_book_bvoc_2013}.

% We propose to tackle our problem as a \gls{misr} task, in which a \gls{hr} isoprene emission map is estimated using multiple \gls{lr} maps, i.e., in our case, an isoprene emissions map and several \gls{lr} land cover maps as suitable emission drivers. This task opposes to the \gls{sisr} task, which only uses a single map as input for the \gls{sr} process~\cite{giganti2023icip}.
%--------------------------------------------------------- 

% NEW VERSION-----------------------------------------------
Accurate \gls{hr} isoprene emission maps are critical for understanding \gls{bvoc} dynamics and their environmental impacts. However, generating such maps at fine spatial resolutions remains challenging due to limitations in observational data and computational constraints in emission modeling. While traditional \gls{sisr} methods can enhance \gls{lr} emission maps using \gls{dl} models~\cite{giganti2023icip, nguyen_sisr_lst_2022, brecht_sisr_wind_2023, izumi_sisr_sst_2022}, these approaches often rely solely on the inherent patterns within the \gls{lr} input, neglecting ancillary environmental variables that drive emission processes~\cite{cai_bvoc_scientometric_2021, guenther_megan3_2020, silibello_bvoc_south_italy_2022}.

In our prior work~\cite{giganti2023icip}, we demonstrated the feasibility of applying \gls{sisr} to isoprene emissions, reconstructing \gls{hr} estimates from \gls{lr} inputs using a \gls{dl}-based \gls{sr} network. However, this single-input formulation inherently limits the model's ability to leverage domain-specific knowledge about the environmental drivers of isoprene emissions, such as \gls{lc} characteristics. \gls{lc} data encode critical ecological and environmental factors (e.g., vegetation type, canopy structure, and soil properties) that strongly influence biogenic emissions, as evidenced by extensive studies~\cite{penuelas_bvocs_2010, stavrakou_isoprene_asia_2014, ciccioli_bvoc_2023, ashworth_book_bvoc_2013, opacka_isoprene_2021, zhang_bvoc_sichuan_2022, bauwens_past_future_bvoc_mohycan_2018, wang_megan32_2024}. Integrating these drivers as auxiliary inputs could improve \gls{sr} accuracy by grounding the reconstruction process in physically meaningful relationships~\cite{dp2024isopreneCMIP6, opacka_isoprene_2021, dramsch_explainability_2025, li_improving_aq_2023}.

In this work, we reformulate the problem as a \gls{misr} task, where an \gls{hr} isoprene emission map is estimated by jointly leveraging (i) the \gls{lr} isoprene emission map and (ii) multiple \gls{lr} \gls{lc} maps representing key emission drivers. This shift from \gls{sisr} to \gls{misr} allows the model to exploit spatial and cross-feature correlations between isoprene emissions and \gls{lc} variables, like vegetation type and urbanization levels. 

% By combining multi-source environmental data with \gls{dl}, our approach aims to produce \gls{hr} emission maps that are spatially refined and more consistent with the underlying ecological processes governing isoprene emissions.

We start considering the isoprene \gls{lr} emission $\LR$ that we want to super-resolve.
Since isoprene emissions exhibit spatially sparse patterns and a wide dynamic range, we apply a non-linear transformation $\T(\cdot)$ to the \gls{lr} isoprene emission map, adopting the approach proposed in~\cite{giganti2023icip}. This transformation is derived from statistical analysis of the available \gls{hr} data and is used to increase the robustness to outliers and local maxima, adapting the emission dynamics into more feasible values required for numerical stability when training \glspl{nn}. We define the transformed emissions as $\HRt$ and $\LRt$ for the \gls{hr} and \gls{lr} version, respectively.

Then, we stack the transformed isoprene emission with $D$ emission drivers, i.e., the \gls{lc} maps. 
In specific, the emission drivers are denoted with the set $\DLR = \{\mathbf{D}_1, \mathbf{D}_2, \dots, \mathbf{D}_D\}$. 
The selected emission drivers cover the same geographical area of the \gls{lr} isoprene map they are stacked with.
In stacking a single \gls{lr} isoprene emission with $D$ additional emission drivers, we obtain $C = 1 + D$ emission maps, represented by the expression $\{ \LRt, \DLR \}$.
%The goal is to super-resolve the isoprene emission $\LR$ via the following approach, adapted from \cite{giganti2023eusipco}.

We propose to estimate an \gls{hr} isoprene emission map as
\begin{equation}
\label{eq:method}
    \HRhat = \Tinv(\Net( \{ \LRt, \DLR \})),
\end{equation}
where $\Net(\cdot)$ is a \gls{nn} architecture specifically designed for \gls{sr} tasks. The operator $\Tinv(\cdot)$ is the inverse transformation of $\T(\cdot)$, needed to re-adapt the super-resolved isoprene emissions to the \gls{bvoc} dynamics~\cite{giganti2023icip}.

% The emission driver domains are in [0,1] and thus do not require transformation.

\begin{figure}[ht]
    \centering
    \includegraphics[width=0.99\linewidth]{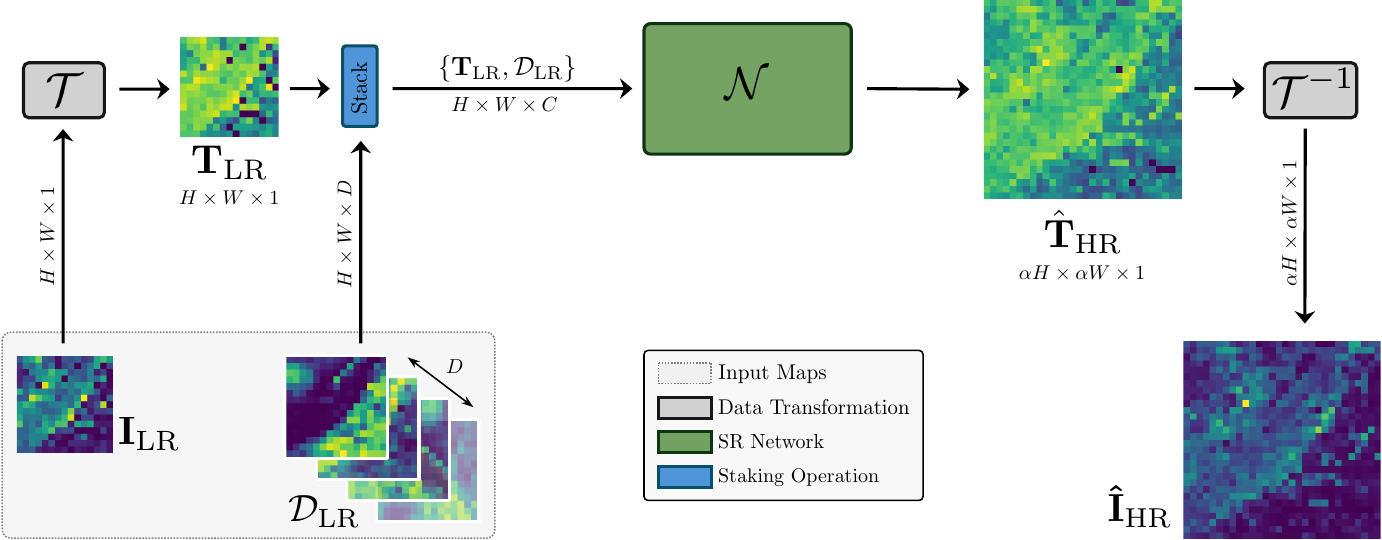}
    \caption{The proposed deployment pipeline of our system for isoprene emission \gls{sr}, leveraging additional emission drivers in the \gls{sr} process.}
    \label{fig:deployment}
\end{figure}

Figure~\ref{fig:deployment} depicts a sketch of the proposed methodology.
We model the stacked set $\{ \LRt, \DLR \}$ as a tensor with a size of $H\times W \times (1 + D)$, where $H$ and $W$ represent the height and width of the maps and $D$ the number of emission drivers exploited in performing \gls{sr}. 
The original $\HR$ and the super-resolved emission $\HRhat$ have size $\alpha H\times \alpha W$, with $\alpha > 1$ indicating the \gls{sr} factor, i.e., how much we increase the spatial resolution of the \gls{lr} isoprene maps. 

In the following, we provide details of the different modules that are included in our pipeline.

\subsubsection{Isoprene Data Transformation}
In our past investigations~\cite{giganti2023icip, giganti2024mdpi}, we showed that a suitable data transformation is required to deal with biogenic emissions, i.e., isoprene, since they are characterized by sparsity, extremely small values, and wide dynamic ranges (from $10^{-18}$ to $10^{3}$ [molec./cm\textsuperscript{2}s]~\cite{isoprene_seeds_2024}).
As a matter of fact, isoprene emissions can present many outliers due to the large spatial diversity of the environmental factors driving the emission process, such as meteorology, type of vegetation, seasonal cycle, and atmospheric composition~\cite{sindelarova_megan21_2022, wang_megan32_2024}. 
For this reason, we adopt a non-parametric data transformation $\T(\cdot)$ that forces emission values to follow a uniform distribution between $0$ and $1$~\cite{scikit-learn, giganti2023icip, peterson_oqn_2020}, enhancing robustness to outliers and local maxima, and at the same time, ensuring numerical stability during neural network training.

\subsubsection{Super-Resolution Neural Networks}
The choice of employing a \gls{nn} as the \gls{sr} operator $\Net(\cdot)$ is driven not only by the demonstrated superior performance of modern \gls{dl}-based methods compared to traditional approaches but also by their inherent adaptability and flexibility. Specifically, in the context of \gls{misr}, \gls{dl}-based solutions offer a significant advantage by eliminating the need to explicitly define a formal relationship between the various emission drivers, like \gls{lc} information related to the geographic area to be super-resolved. This means that they can effectively capture the intricate dependencies between multiple diverse types of input images, thereby enhancing the overall performance and robustness of the \gls{sr} task.

% Instead, \gls{dl}-based approaches excel by leveraging their ability to uncover and utilize any advantageous connections or patterns embedded within the input data during the training process. This is particularly beneficial because such relationships can be highly complex or even unknown. 

We investigate two \glspl{nn} that are the state-of-the-arte in \gls{sr}, each leveraging advanced attention mechanisms: (i) the \gls{hat}~\cite{chen_hat_2023}, which combines transformer-based global attention~\cite{bishop_dl_book_2024} with local detail refinement; (ii) the \gls{san}~\cite{dai_sansisr_2019}, which utilizes second-order feature interactions to enhance the reconstruction of complex spatial patterns.

Both networks address the challenge of modeling long-range dependencies. \gls{hat} uses \glspl{ocab}~\cite{chen_hat_2023} to effectively aggregate multi-scale contextual information by overlapping patches during attention computation. In contrast, \gls{san} incorporates \gls{rlnl} modules~\cite{dai_sansisr_2019} that operate on smaller image regions. The choice of these mechanisms could influence the network's effectiveness in handling complex patterns and textures that span more significant emission map regions.

Notably, \gls{san} has already been successfully applied to the \gls{sr} of biogenic emissions~\cite{giganti2023icip, giganti2023eusipco, giganti2023igarss, giganti2024mdpi}. Transformer-based \gls{sr} networks for climate data are less explored in the literature, though the few existing applications show promising results~\cite{carbone_sr_sentinel_2024}.

\subsection{Training Pipeline}
\begin{figure}[ht]
    \centering
    \includegraphics[width=0.99\linewidth]{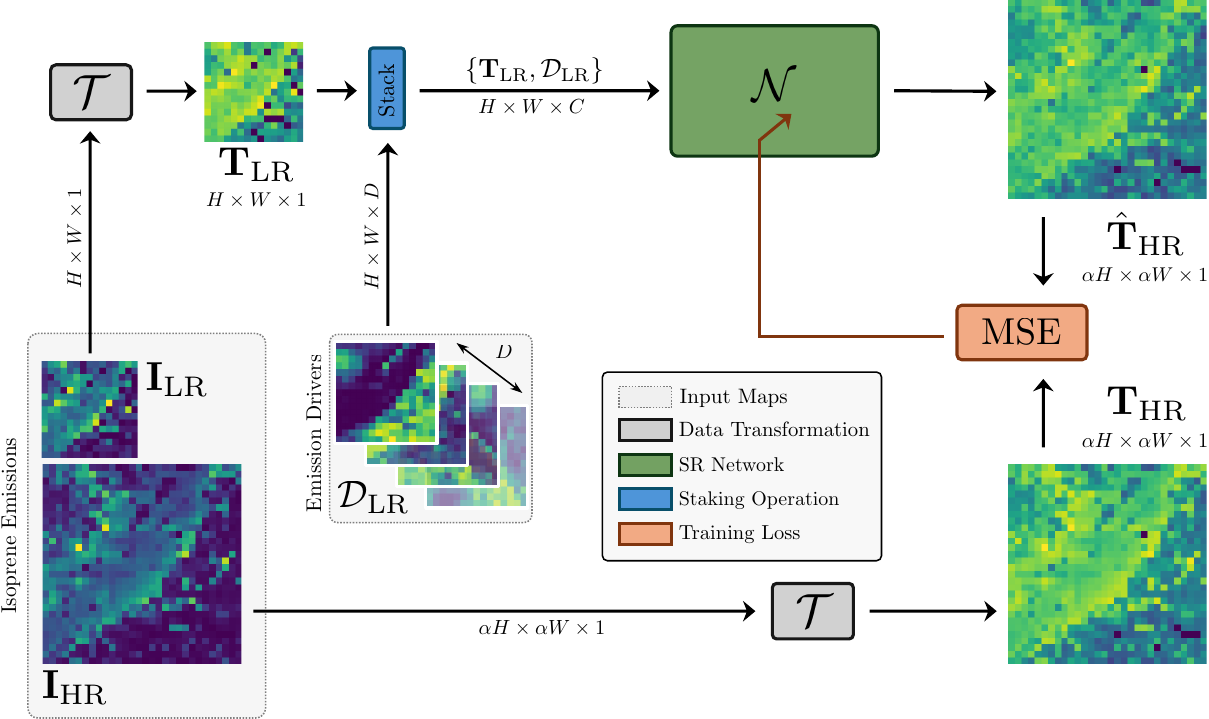}
    \caption{The proposed training pipeline of our system for isoprene emission \gls{sr}, leveraging additional emission drivers in the \gls{sr} process.}
    \label{fig:training}
\end{figure}
We present the proposed training pipeline in Figure~\ref{fig:training}.
We begin by considering the \gls{hr} and its corresponding \gls{lr} version of isoprene emissions, denoted as $\HR$ and $\LR$, respectively.
The selected $D$ emission drivers from the $\DLR$ set are stacked with a transformed version of $\LR$, i.e., $\LRt = \T(\LR)$.
Thus, after the stacking operation, we obtain a tensor $\{ \LRt, \DLR \}$ with dimensions $H \times W \times C$, where $C = 1 + D$.

This tensor is then fed into our \gls{sr} network $\Net$, which outputs a super-resolved version of the isoprene map $\HRhatt$, with dimensions $\alpha H \times \alpha W \times 1$, where $\alpha$ is the scale factor that represents the increase in spatial resolution.

We use the \gls{mse} as the loss function between the super-resolved map $\HRhatt$ and the ground truth $\HRt = \T(\HR)$.

%The data transformation $\T$ instead is estimated following the approach proposed in~\cite{giganti2023icip}, therefore considering only information from \gls{hr} isoprene emission from the training partition.

% In the deployment phase, reported in Figure~\ref{fig:deployment}, given a \gls{lr} isoprene emission map $\LR$ and a set of $D$ \gls{lr} emission drivers in $\DLR$, we estimate an \gls{hr} isoprene emission map $\HRhat$. 

\section{Experimental Dataset}
\label{sec:experimental_dataset}
\subsection{Study Area}
\begin{figure}[ht]
    \centering
    \includegraphics[width=0.99\linewidth]{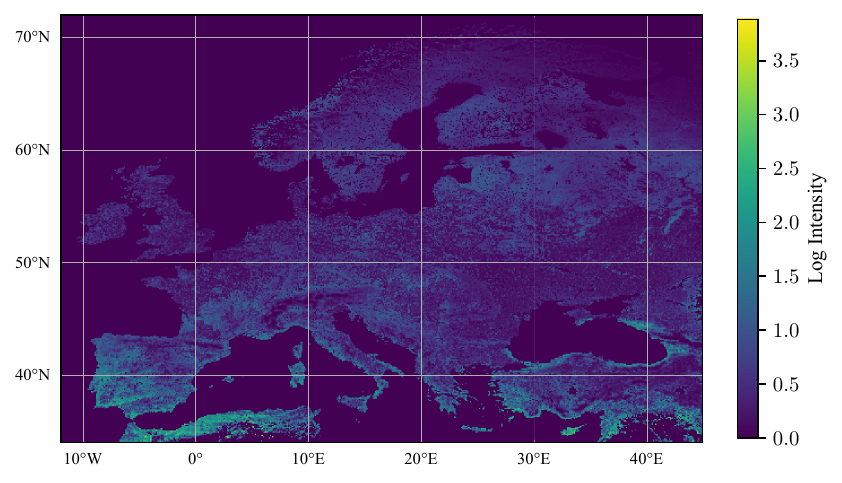}
    \caption{The study area adopted in this work; the map report a monthly averaged isoprene emission (April 2019) from the $\seeds$ inventory. Emission is reported as $\frac{mol}{cm^{2}s{1}}$.}
    \label{fig:study_area}
\end{figure}

% Talk about SEEDS, report the study area from SEEDS
We select the European region as our study area, specifically focusing on the geographical extent defined by the $\seeds$ isoprene inventory, i.e., 11.95° W, 44.95° E, 34.05° N, 71.95° N.
Figure~\ref{fig:study_area} provides a detailed illustration of the study area.

\subsection{Pre-Processing Workflow}
\subsubsection{Land Cover Maps}
We aim to super-resolve isoprene emission maps obtained from the $\seeds$ inventory by exploiting \gls{lc} information from the $\esawc$. 
As presented in Section~\ref{ssec:land_cover}, the $\esawc$ dataset only captures a single temporal state (2021 snapshot), but since \gls{lc} changes occur relatively slowly~\cite{opacka_isoprene_2021}, it is assumed to be valid over the full observation period of $\seeds$.

The $\esawc$ dataset features 11 \gls{lc} classes. We extract two different \gls{lc} types relevant to biogenic emissions. In specific, the \gls{cl} and \gls{tc} maps are selected since they represent very informative biogenic emission drivers, as reported in Section~\ref{ssec:statistical_relationships}.
%To provide a more nuanced view of land cover distribution
Instead of representing discrete classes (binary or categorical), we compute the proportion of each land type within the area represented by each grid cell. This transformation results in two distinct percentage maps, where each cell contains a continuous value between 0\% and 100\%.
These maps provide a detailed view of the selected isoprene emission drivers, capturing mixed land uses within cells and enabling analysis of gradients, patterns, and changes while supporting \gls{sr} models with continuous data.

\begin{figure*}[ht]
    \centering
        \begin{subfigure}{0.3247\textwidth}
        \centering
        \includegraphics[width=\linewidth]{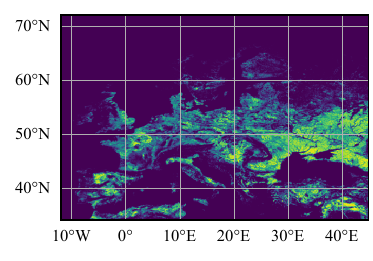}
        \caption{\gls{cl}}
        \label{fig:cl_example}
    \end{subfigure}
    \hfill
    \begin{subfigure}{0.341\textwidth}
        \centering
        \includegraphics[width=\linewidth]{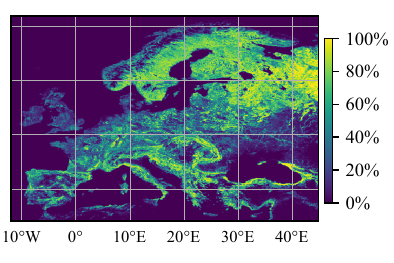}
        \caption{\gls{tc}}
        \label{fig:tc_example}
    \end{subfigure}
    \hfill
    \begin{subfigure}{0.3121\textwidth}
        \centering
        \includegraphics[width=\linewidth]{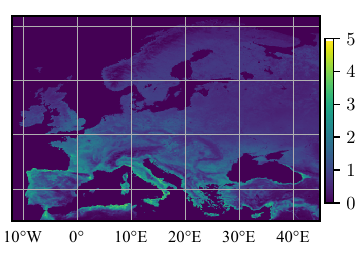}
        \caption{\gls{lai}}
        \label{fig:lai_example}
    \end{subfigure}
    \caption{The \gls{cl} (\ref{fig:cl_example}), \gls{tc} (\ref{fig:tc_example}) percentage maps and \gls{lai} information (\ref{fig:lai_example}) over the study area from the $\esawc$ and $\laiseeds$ inventories, respectively. For the \gls{lai} map in (\ref{fig:lai_example}), we report a monthly averaged value (April 2019).}
    \label{fig:cl_tc_lai}
\end{figure*}

After that, we spatially confine the two percentage maps from $\esawc$ to the study area of the $\seeds$. Due to the different spatial resolutions among the two datasets, a cell coordinate alignment process followed by spatial resolution matching is required. This is obtained by applying a coarsening function~\cite{hoyer2017xarray} over the entire map. % with a factor of 0.75. 
After that, a bicubic interpolation function is used to align the resampled data to $\seeds$ coordinates, ensuring that both datasets are consistently aligned, providing a congruous input of \gls{lc} to isoprene emission data for \gls{sr} model across the study area.
Figure~\ref{fig:cl_example} and Figure~\ref{fig:tc_example} report the \gls{cl} and \gls{tc} \gls{lc} types, respectively, as percentages of the surface area underlying each grid cell over the full study area.

\subsubsection{Leaf Area Index Maps}
In addition to the \gls{lc} maps, we also investigate the adoption of \gls{lai} information as an additional prior for the isoprene \gls{sr} task. As reported in Section~\ref{ssec:lai}, the $\laiseeds$ inventory~\cite{lai_seeds_2024} provides the total leaf area per unit ground area. Similar to isoprene emissions, these maps cover Europe from 2018 to 2022, with a spatial resolution of $0.1\degree \times 0.1\degree$. In this inventory, \gls{lai} values range from $4.17 \times 10^{-5}$ to $7.2$. Although \gls{lai} is defined as the ratio of the number of square meters of leaves per square meter of land surface, i.e., $\frac{m^2}{m^2}$, values greater than one indicate that the area is covered with multiple layers of leaves, as in a forest canopy, where vegetation grows in layers above the ground. For instance, an \gls{lai} value of 5 conceptually represents five layers of leaves in that area. In Figure~\ref{fig:lai_example} we report the monthly averaged \gls{lai} map (April 2019) over the entire study area.

\subsubsection{Climate Class Maps}
\begin{figure}[ht]
    \centering
    \includegraphics[width=0.99\linewidth]{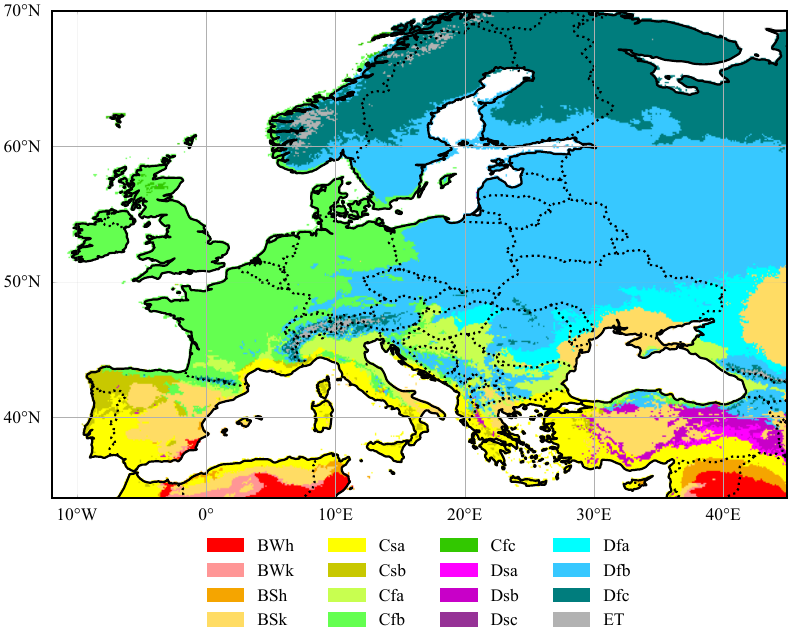}
    \caption{Geographical distribution of the Köppen-Geiger climate classes that are present over the study area. Please refer to Table~\ref{tab:kg_classes_study_area} for more information regarding the different climate classes.}
    \label{fig:kg_study_area}
\end{figure}

For a more reliable evaluation of the proposed methodology, we analyze the \gls{sr} performance over the different Köppen-Geiger climate classes present in our study area. 
The climate classes are extracted from the $\kgclim$ dataset. As presented in Section~\ref{ssec:climate_zones}, the $\kgclim$ dataset contains six different static maps, which differ in the years they cover. We adopt the map covering the 1991–2020 period, as this is the period of most significant overlap with the isoprene emissions in the $\seeds$ inventory.
We choose the native $0.1\degree \times 0.1\degree$ resolution of the $\kgclim$ to match the spatial resolution of the $\seeds$ inventory.
As previously done for the \gls{lc} maps from the $\esawc$ dataset, the coordinates of the climate class maps are aligned, and the spatial extent is reduced to match the study area set by the $\seeds$ inventory.
Figure~\ref{fig:kg_study_area} reports the geographical distribution of the Köppen-Geiger climate classes over the entire study area.
In addition, Table~\ref{tab:kg_classes_study_area} report a detailed description for each climate class type~\cite{beck_kg_climate_2023}.

\begin{table}[t]
    \caption{Overview of the Köppen-Geiger climate classes that are present over the study area. Please refer to the original work in~\cite{beck_kg_climate_2023} for the defining criteria of each class.}
    \centering
    \resizebox{0.9\columnwidth}{!}{ 
    \begin{tabular}{ll}
        \toprule
        \textbf{Climate Class} & \textbf{Description} \\
        \midrule
        BWh & Arid, desert, hot \\
        BWk & Arid, desert, cold \\
        BSh & Arid, steppe, hot \\
        BSk & Arid, steppe, cold \\
        Csa & Temperate, dry summer, hot summer \\
        Csb & Temperate, dry summer, warm summer \\
        Cfa & Temperate, no dry season, hot summer \\
        Cfb & Temperate, no dry season, warm summer \\
        Cfc & Temperate, no dry season, cold summer \\
        Dsa & Cold, dry summer, hot summer \\
        Dsb & Cold, dry summer, warm summer \\
        Dsc & Cold, dry summer, cold summer \\
        Dfa & Cold, no dry season, hot summer \\
        Dfb & Cold, no dry season, warm summer \\
        Dfc & Cold, no dry season, cold summer \\
        ET & Polar, tundra \\
        \bottomrule
    \end{tabular}
    }
    \label{tab:kg_classes_study_area}
\end{table}

% As presented in Section~\ref{sec:dataset}, the $\esawc$ dataset only captures a single temporal state (2021 snapshot), but since land cover changes occur relatively slowly [CITE], it is assumed to be valid over the full observation period of $\seeds$.
% The $\esawc$ dataset features 11 land cover classes. We extract two different land cover types relevant to BVOC emissions, i.e., cropland and tree cover, since these represent informative biogenic emission drivers, as reported in Section~\ref{ssec:preliminary_results}.

\subsection{Patch Extraction}
\label{ssec:patch_extraction}
% -- High-Resolution Patch Extraction
Considering the $\seeds$ inventory extension, we extract multiple \gls{hr} patches from the entire maps with a $3\degree\times 3\degree$ spatial coverage corresponding to $30 \times 30$ grid cells. 
%Each patch is defined by a bounding box with minimum and maximum longitude and latitude coordinates.
Patches are created with a $1\degree$ stride to ensure continuous spatial coverage. Therefore, a $2\degree$ overlap in latitude and longitude over the study area is considered.
%, extending from $11.95\degree W to 44.95\degree E$ in longitude and $34.05\degree N to 71.05\degree N$ in latitude.

% -- isoprene Emission Filtering
Following previous works on isoprene \gls{sr}~\cite{giganti2023icip, giganti2023igarss, giganti2024mdpi}, a zero-thresholding criterion on isoprene emission patches is applied, promoting stability in the \gls{sr} network training. Therefore, patches with more than 10\% zero-value emission fluxes are discarded, as they provide insufficient isoprene information. This process results in a total number of $913,878$ isoprene emission patches.

% -- Dataset Index Creation
For an extensive experimental evaluation, patch-specific information, like the coordinates boundaries, daily time, and climate classes, are extracted, resulting in a complete dataset index that associates each patch with its specifications across the European extent of $\seeds$.

% -- Low-Resolution Patch Creation
From the \gls{hr} isoprene emission maps and their associated emission drivers maps, we generate their \gls{lr} counterpart by performing bicubic downsampling, obtaining maps of $15\times15$ grid cells.
Our goal is to estimate \gls{hr} isoprene emission patches $\HR$ with $0.1\degree\times 0.1\degree$ spatial resolution starting from their \gls{lr} counterparts $\LR$ and their related drivers in $\DLR$, both having a $0.2\degree\times 0.2\degree$ spatial resolution, thus addressing a \gls{sr} scale factor $\alpha=2$. 

The extracted patches related to isoprene and its emission drivers form the complete experimental dataset. We denote the experimental dataset as $\AllSet$.

\subsection{Climate-specific Folding}
\label{ssec:climate-specific_folding}
Climate variability plays a crucial role in shaping isoprene emissions' spatial and temporal distribution. Different climate zones exhibit distinct interactions between meteorological conditions, vegetation types, and emission drivers, making it challenging to develop a single model that generalizes well across all regions. 

To address this, we adopt a specialized experimental strategy.
From the entire experimental dataset $\AllSet$, we generate different folds 
%$\{\Foldn\}$, 
covering the same temporal period, i.e., 2018-2022. 
Each fold, denoted as $\Foldn$, is structured around a specific climate class (\textit{cc}) that is held-out from the experimental data.  Patches from the excluded climate class are treated as representing an unseen climate zone, enabling subsequent generalization experiments in Section~\ref{sssec:climate_scenario}.

Based on the distribution of climate classes across the study area, the three most frequently occurring classes, along with one custom Köppen-Geiger climate class, are selected as the held-out climate classes.

The selected classes are $\Climone$, $\Climtwo$ and $\Climthree$.
In addition, we define a custom climate class $\Climfour$, where we include minor and sporadically occurring climate classes that are present in the Mediterranean area, making sure to reach a similar number of patches of the other climate classes.
Creating a custom climate class was necessary because the geographical areas covered by the climate classes in $\Climfour$ represent the study area under consideration. However, when taken individually, these classes do not contain sufficient patches to be considered a meaningful set for subsequent studies. For these reasons, we merged a subset of relevant climate classes for the study area to characterize it better.

% The \gls{medfld} encompasses various climate classes present in our study area and proved to be a suitable set for the overall performance of our proposed methodology.

Therefore, from the experimental dataset $\AllSet$, we generate four different folds, specifically $\{\Foldone, \Foldtwo, \Foldthree, \Foldfour\}$. Each fold excludes patches belonging to a selected climate class, i.e., $\Climone$, $\Climtwo$, $\Climthree$ and $\Climfour$.

\begin{figure}[ht]
    \centering
    \includegraphics[width=0.99\linewidth]{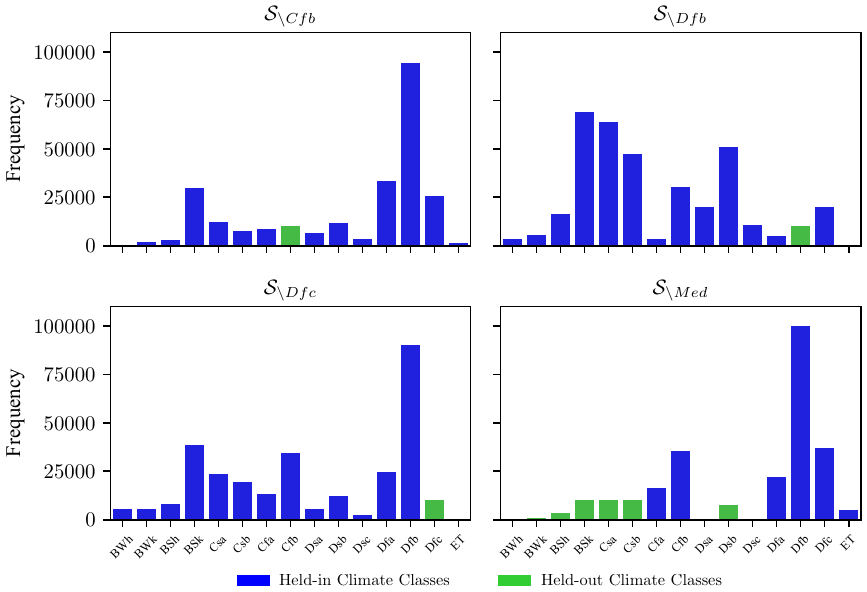}
    \caption{The considered Köppen-Geiger climate classes for each $\Foldn$ folds and their occurrences (number of patches) over the study area.}
    \label{fig:climate_holdout_classes}
\end{figure}

Figure~\ref{fig:climate_holdout_classes} reports the occurrences of the Köppen-Geiger climate classes over the study area. Colors denote which climate classes are considered for each of the four different dataset folds $\Foldn$.

\begin{figure}[t]
    \centering
    \includegraphics[width=0.99\linewidth]{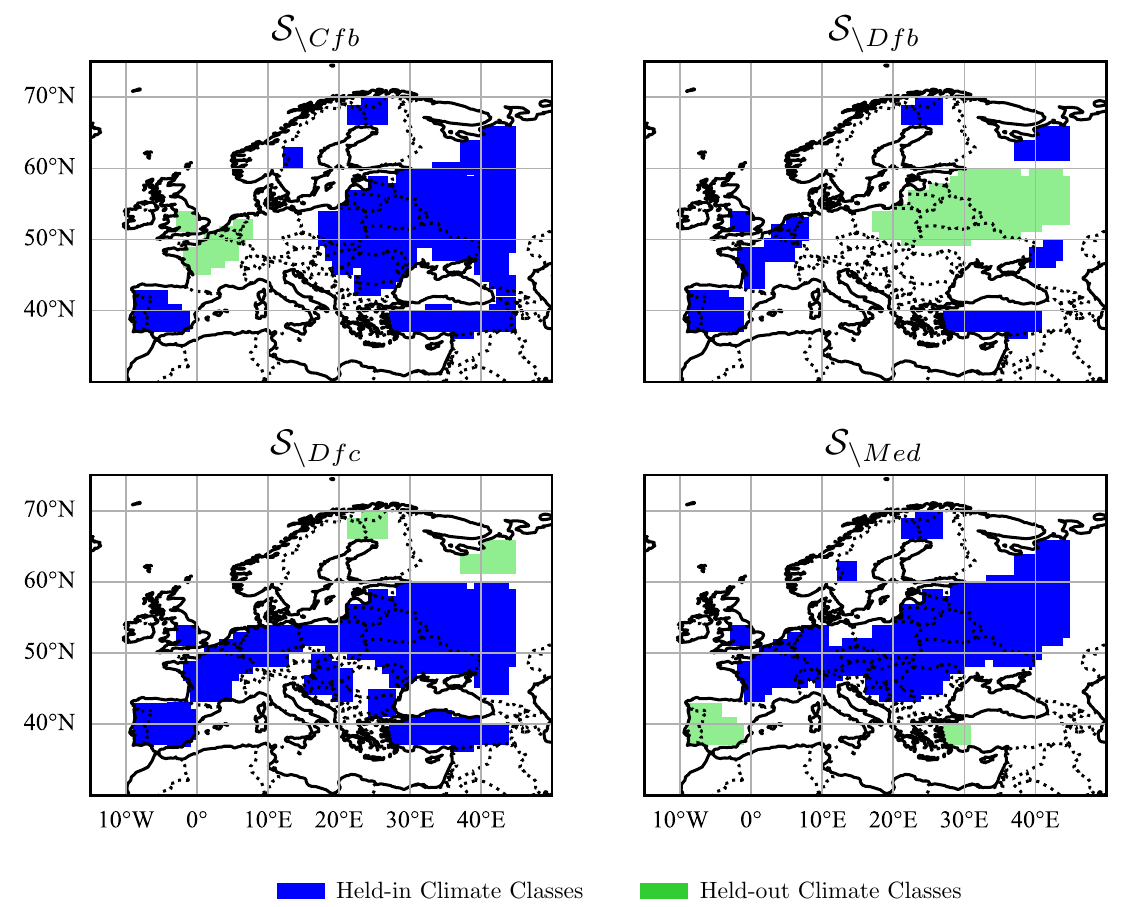}
    \caption{Geographical distribution of patches for each fold $\Foldn$.}
    \label{fig:standard_scenario_geog_distr}
\end{figure}

In Figure~\ref{fig:standard_scenario_geog_distr} instead, we report the geographical distribution of the 
%selected patches for each of the 
four climate-specific folds $\Foldn$.

% Therefore, we train four models, one for each $\Foldn$ fold, thus excluding patches from one specific climate class, rather than training a separate model for each one. 
% % This approach allows us to assess the model's ability to generalize to unseen climate conditions while leveraging shared patterns across similar climate zones. 
% In this way, the model is provided with diverse yet relevant training data, ensuring robust learning by leveraging shared emission patterns across similar climate zones. 
%while having a set of patches for testing its extrapolation capabilities in unseen climate conditions.

\subsection{Experimental Scenarios}
\label{ssec:experimental_scenarios}
To evaluate the capability of our proposed methodology, for each climate-specific fold $\Foldn$ we analyze the model's performance under three different scenarios:
\begin{enumerate}
    \item \textit{Standard scenario}, where training and evaluation data are coherent in terms of climate zones and geographical areas.
    \item \textit{Unseen spatial areas scenario}, in which we evaluate our method over geographical areas never seen in training phase.
    \item \textit{Unseen climate zone scenario}, where we test over climate zones never seen in training phase. 
\end{enumerate} 
The purpose of these proposed scenarios is to compare and assess the model's ability to super-resolve emissions with different environmental characteristics.
Below, we provide additional details and motivations for these experimental scenarios, highlighting their significance in addressing climate-specific challenges and improving the model's generalization capabilities across diverse environmental conditions.

\subsubsection{Standard Scenario}
\label{sssec:standard_scenario}
%In this scenario, we train and test four different models, each considering emissions from a single climate-specific fold $\Foldn$ of the complete dataset $\AllSet$, as presented in~\ref{ssec:climate-specific_folding}.

In this scenario, 
%for each climate-specific fold $\Foldn$, we begin by randomly sampling $100,000$ patches from the selected fold $\Foldn$.
%These patches are then 
for each fold $\Foldn$, we randomly split the held-in patches using a $75/5/20\%$ strategy for the training, validation, and test partitions, respectively.
In this way, the test partition contains patches that share the same geographical and climate coverage of the train and validation partitions.
Referring to fig.~\ref{fig:standard_scenario_geog_distr}, this scenario considers only patches randomly extracted from the 
% Figure~\ref{fig:holdout_geog_distr} 
% reports the geographical distribution of patches per fold. Specifically, 
blue areas for each climate fold.
%are referred to the patches associated with the standard scenario.

\subsubsection{Unseen Spatial Areas Scenario}
\label{sssec:spatial_scenario}
This scenario focuses on investigating \gls{sr} generalization across different geographical areas. This aspect is of great significance~\cite{yadav_deepaq_2023, medina_xai_dl_ctm_2024}, especially given the comparatively strong performance of temporal generalization, as noted in previous studies on isoprene emission \gls{sr}~\cite{giganti2023icip}.

To achieve this, for each fold $\Foldn$, we create a spatial hold-out subset containing $10,000$ patches from a specific area within the geographical extent of the climate-specific fold $\Foldn$. This 
%Therefore, after selecting a geographical point inside the geographical extent of $\Foldn$, we identify the  geographically closest patches to that point, forming the 
spatial hold-out subset is then used as a test set for our model.
Notice that the 
selected patches 
%span various climate classes, including some already present in the training dataset, 
potentially include all the climate zones seen in training phase,
allowing the \gls{sr} models to be tested on unseen spatial features while remaining within a familiar climate context. 
Figure~\ref{fig:holdout_geog_distr} 
reports the geographical distribution of patches per fold. Specifically, the red area refers to the patches associated with the unseen spatial areas scenario.

\subsubsection{Unseen Climate Zones Scenario}
\label{sssec:climate_scenario}
In this last scenario, we analyze the performance of our method by investigating the \gls{sr} generalization across different climate conditions, testing the \gls{sr} model on emission patches with unseen climate during training.

\begin{figure}[ht]
    \centering
    \includegraphics[width=0.99\linewidth]{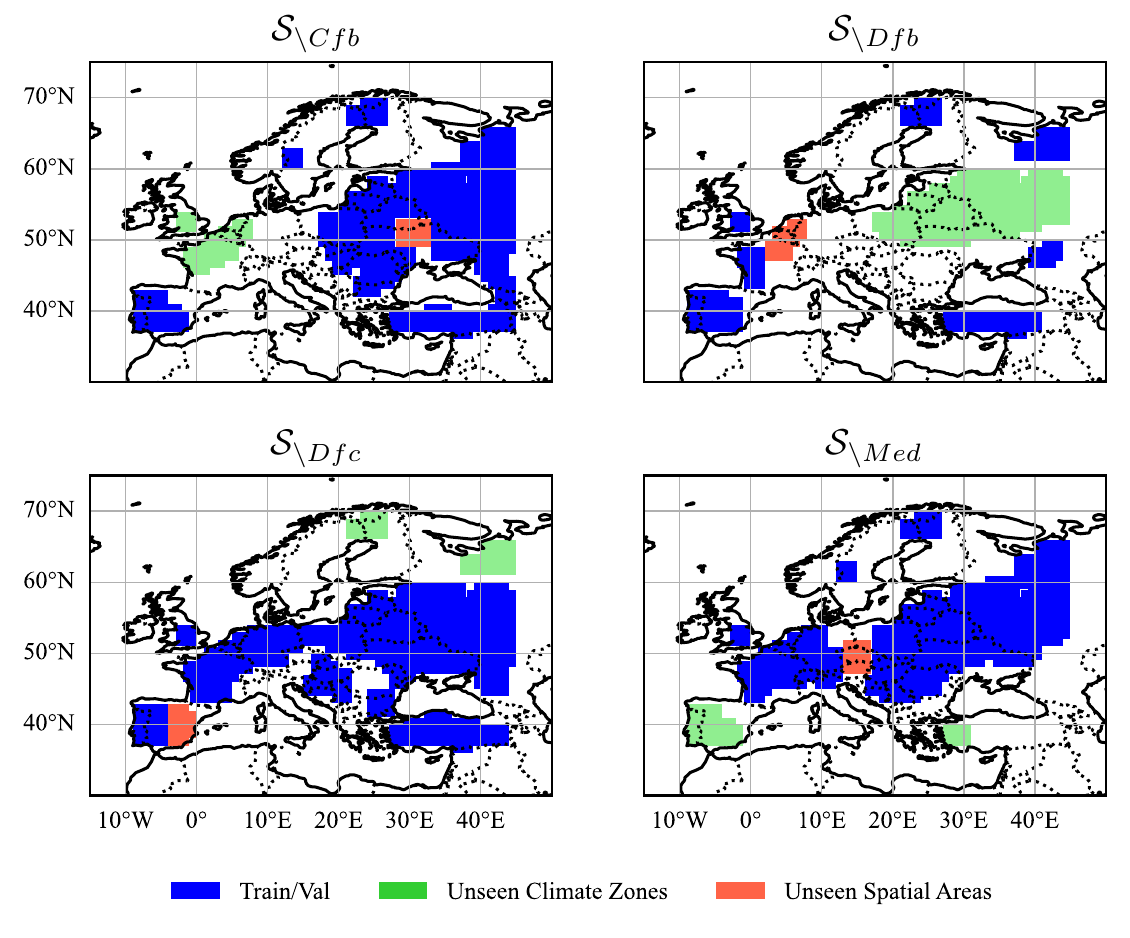}
    \caption{Geographical distribution of patches for each fold $\Foldn$, associated with the standard (train and validation patches), unseen spatial areas and unseen climate zones scenarios.}
    \label{fig:holdout_geog_distr}
\end{figure}

Since each fold $\Foldn$ is structured in a way that a specific climate class (\textit{cc}) is held out from the training data (see Section~\ref{ssec:climate-specific_folding} for more details), some patches from that held out climate class are selected as our test set.
%the climate hold-out subset that is used here as the test set.
Specifically, for each fold $\Foldn$, $10,000$ patches from the held out climate class (\textit{cc}) are randomly selected for testing. 
Figure~\ref{fig:holdout_geog_distr} 
reports the geographical distribution of patches per fold. Specifically, green areas refer to patches associated with the unseen climate zone scenario.
%the \gls{sr} model on regions with climate they have not encountered during training, leveraging the dataset's climate diversity to assess generalization.

% \begin{figure}[ht]
%     \centering
%     \includegraphics[width=0.9\linewidth]{imgs/Partition_Time_Dist.pdf}
%     \caption{Spatiotemporal distribution for each fold $\Foldn$, associated with the Standard (train and validation patches), Climate and Spatial scenarios. Patches are sorted first by ascending latitude, then ascending time per latitude (from left to right).}
%     \label{fig:holdout_temporal_distr}
% \end{figure}

\vspace{10pt}
The unseen areas and climate zone scenarios aim to ensure that each fold tests the performance of the \gls{sr} model on distinct climate classes and geographical locations. 
Therefore, with this approach, we aim to capture most of the dataset's variance to assess the \gls{sr} performance accurately.

\section{Experimental Setup}
\label{sec:experimental_setup}
\subsection{Training Settings}
\label{ssec:training}
To estimate the data transformation $\T(\cdot)$, we follow the experimental setup proposed in \cite{giganti2023icip}.
To train the \gls{sr} networks we use Adam as optimizer with $\beta_1 = 0.9$, $\beta_2 = 0.99$, and an initial learning rate of $10^{-4}$. We decrease the learning rate by a factor $10$ if the validation loss does not improve in 10 epochs until the minimum learning rate $10^{-7}$ is reached. 
We stop the learning phase if the validation loss does not improve after $50$ epochs, setting a maximum number of epochs of $500$.
%We use a batch size of $200$ elements.
%The \gls{hat} was trained for up to $500$ epochs, 
%(except for \gls{26fld}, where early stopping determined the required epochs for the other runs),
% while the \gls{san} was limited to $150$ epoch.
%Each batch element is a pair of tensors $\{ \LRt, \DLR \}$ with size of $H\times W \times C$, with $C = D + 1$, and the transformed isoprene emission $\HRt$ with size of $\alpha H\times \alpha W \times 1$.
%A simplified version of our training setup is shown in Figure~\ref{fig:training}.

In our experiments, we use an NVIDIA Titan V 12 GiB GPU (with 5120 CUDA cores at 1455MHz) running on an Intel Xeon E5-2687W v4 CPU (with 48 Cores at 3 GHz) equipped with 252 GiB of RAM.

Kindly refer to the released implementation code for additional training and network details.

\subsection{Evaluation Metrics}
\label{ssec:metrics}
For a quantitative evaluation of the proposed method effectiveness, we conduct a comparison between the original \gls{hr} isoprene emissions $\HR$ and their super-resolved versions $\HRhat$ by adopting several metrics commonly used in the \gls{sr} literature~\cite{sdraka_dl_sr_rs_2022, li_sr_2023, donini_sr_radargrams_2024, carbone_sr_sentinel_2024}.
We evaluate our method in terms of \gls{ssim}~\cite{ssim_metric}, \gls{psnr}, \gls{uiqi}~\cite{uiqi_metric}, \gls{scc}~\cite{scc_metric}.
In addition, we propose two additional metrics, the \gls{nmse} in dB and \gls{maxae}.

We define the \gls{nmse} as the ratio of the power of the error introduced by the \gls{sr} process to the power of the original image and is defined as
\begin{equation}
    \operatorname{NMSE}(\HRhat, \HR) = 10\log_{10} \left( \frac{\mse(\HRhat, \HR)}{\operatorname{Avg}(\HR^2)} \right)
\end{equation}
where $\operatorname{Avg}$ extract the mean value of the image.

The \gls{maxae} quantifies the worst-case discrepancy between the super-resolved image $\HRhat$ and ground-truth $\HR$ by identifying the maximum absolute difference at any pixel location of a 2D image, and it is defined as 
\begin{equation}
    \operatorname{MaxAE}(\HRhat, \HR) = \max_{m,n} \left( \left| \HRhat(m, n) - \HR(m, n) \right| \right),
\end{equation}
where $m$ and $n$ denote the pixel coordinates of the considered emission map.
Therefore, \gls{maxae} measures the most significant error in the super-resolved image $\HRhat$ compared to the ground truth $\HR$.

The best values for these metrics are 1 for \gls{ssim}, \gls{uiqi}, and \gls{scc}; the higher, the better for \gls{psnr}; the less, the better for \gls{nmse} and \gls{maxae}.

\section{Results}
\label{sec:results}
% In this section, we first motivate the adoption of land cover products as priors for isoprene \gls{sr} by performing a statistical analysis. Then, we demonstrate the performance of our proposed methodology in super-resolving isoprene maps by leveraging additional drivers in the process. Finally, we report the generalization studies related to the hold-out sets.
\glsreset{lai}
\glsreset{cl}
\glsreset{tc}

\subsection{Selecting the Emission Drivers}
\label{ssec:statistical_relationships}
In this section, we conduct a comprehensive statistical analysis to evaluate the strength and significance of the relationships between isoprene emissions and the \gls{lai}, \gls{cl}, and \gls{tc} information. This analysis aims to identify which of these drivers has the most substantial influence on isoprene emissions, providing insights into the key factors governing these emissions.
% The inclusion of the \gls{lai} data in addition to the \gls{lc} information is motivated by the isoprene emission mechanisms~\cite{guenther_megan3_2020, guenther_model_2012, dp2024isopreneCMIP6}, as explained in Section~\ref{sec:introduction}.
%and thereby a hypothesized augmenting relation to isoprene emission data. These preliminary analyses' findings informed the main study's methodological direction.
To assess the relationships across the temporal and spatial extent of the $\seeds$ inventory, we use the \gls{pcc} and the entropy, denoted with $\Entr$.

\begin{figure}[t]
    \centering
    \includegraphics[width=0.9\linewidth]{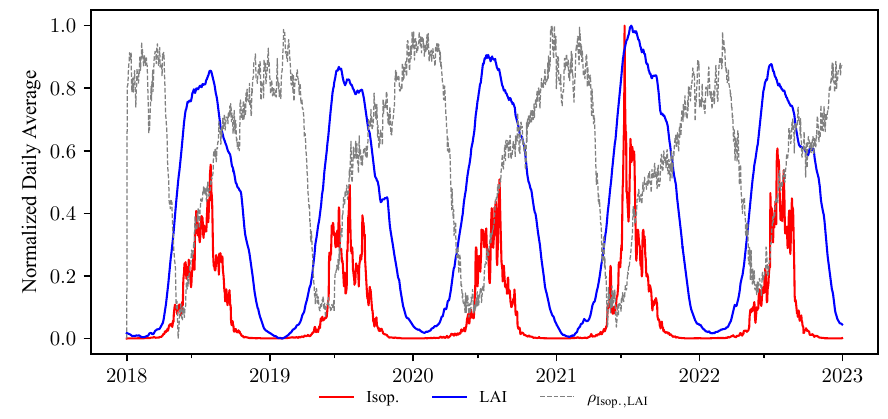}
    \caption{Normalized temporal evolution of \gls{lai} and isoprene intensity (patch averaged), and \gls{pcc} between isoprene and \gls{lai} intensity (patch averaged).}
    \label{fig:pcc_temporal}
\end{figure}

We start by analyzing the temporal dependencies of isoprene emissions and \gls{lai} information. 
\gls{lai} patches from the $\laiseeds$ inventory are extracted following the same operations done for isoprene explained in Section~\ref{ssec:patch_extraction} and have the same geographical extent of our study area. 
% In this analysis, we use the \gls{hr} version of them.
Figure~\ref{fig:pcc_temporal} reports their temporal evolution, considering the mean values of patches, together with the evolution of their \gls{pcc}.
%each patch's mean value. 
%The \gls{pcc} is chosen to inform about the relationship between data on a cell-to-cell basis over time.
Although \gls{lai} mean intensities follow a seasonal cycle similar to that of isoprene emissions, the correlation between isoprene (isop.) emissions and \gls{lai} values (i.e., $\rho_{\mathrm{isop.}, \mathrm{LAI}}$) varies significantly over time. Interestingly, this temporal correlation displays a cyclical pattern offset by half a year from the seasonal cycles of both variables. 
In specific, the correlation peaks are in the winter months, when both \gls{lai} and isoprene emission intensities are at their lowest. 
This finding contrasts with the hypothesis that the \gls{lai}, representing a key variable in the biogenic emissions mechanism, 
%would correspond to and thereby 
could help predict structures within isoprene emission patterns. Therefore, we believe that the \gls{lai} may not be a straightforward proxy for spatial patterns underlying isoprene emission imagery, particularly during periods of high biological activity.

\begin{figure}
    \centering
    \includegraphics[width=0.9\linewidth]{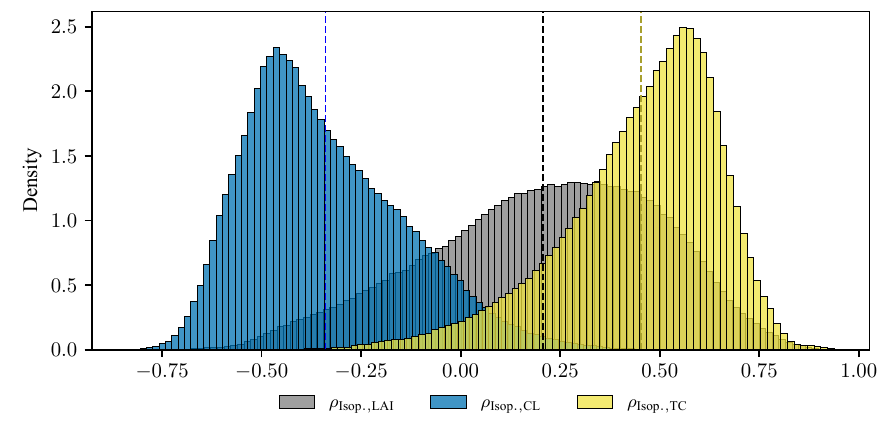}
    \caption{Spatial correlation, measured using the \gls{pcc},  of isoprene emissions with \gls{lai} ($\rho_{\mathrm{isop.}, \mathrm{LAI}}$), \gls{cl} ($\rho_{\mathrm{isop.}, \mathrm{CL}}$) and \gls{tc} ($\rho_{\mathrm{isop.}, \mathrm{TC}}$) information. Vertical dashed lines denote the mean value $\mu$ of the related distribution.}
    \label{fig:pcc_spatial}
\end{figure}

In addition, we analyze the spatial correlation of isoprene emissions with \gls{lai}, \gls{cl}, and \gls{tc} information.
In Figure~\ref{fig:pcc_spatial}, we report the results of this analysis. 
The correlation between isoprene emission and \gls{lai} ($\rho_{\mathrm{isop.}, \mathrm{LAI}}$) is widely distributed (the standard deviation $\sigma = 0.29$) with a low average value of $\mu = 0.21$. 
In contrast, the correlation with \gls{cl} data exhibits a narrower distribution ($\sigma = 0.20$) with a negative $\mu = -0.34$. Lastly, the \gls{tc} correlation also shows a relatively narrow distribution ($\sigma = 0.21$) but has the highest value $\mu = 0.45$. 
Given these results, the 
%This analysis shows that 
\gls{tc} information seems the most suitable candidate out of the three for augmenting isoprene emissions in \gls{sr} applications.
%even though it is not highly correlated.

However, as found in previous works \cite{giganti2023eusipco, lloyd_misr_optically_2022}, we believe that information with inherent negative spatial correlation could be helpful in spatially enhancing isoprene emission maps, suggesting that \gls{cl} data may also be a valuable prior for \gls{sr} purposes. Since the aim is to establish a consistent relationship from additional channels (drivers) to support \gls{sr}, the temporally static nature of \gls{lc} data (single 2021 snapshot \cite{esa_worldcover_v2_2021}) with a relatively narrow distribution of patch correlations makes \gls{cl} and \gls{tc} data potentially suitable for this purpose. 
% This consistency may provide a stable reference, augmenting isoprene emissions data for \gls{sr} applications.

\begin{figure}
    \centering
    \includegraphics[width=0.9\linewidth]{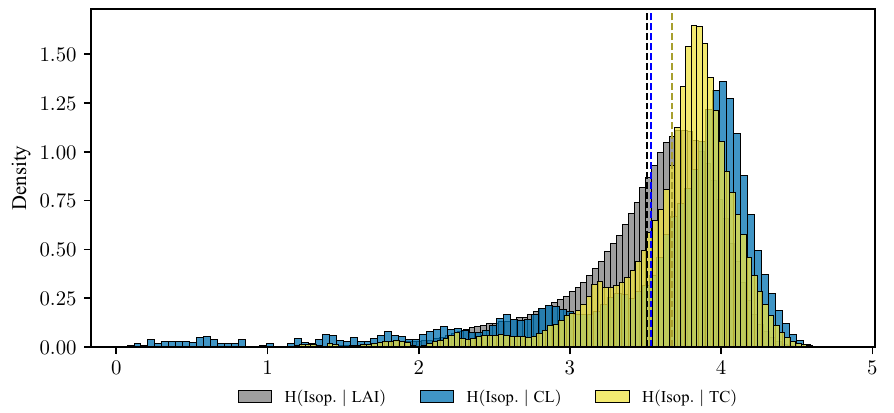}
    \caption{Conditional entropy of isoprene emissions with \gls{lai} ($\Entr(\mathrm{isop.}\ |\ \mathrm{LAI})$), \gls{cl} ($\Entr(\mathrm{isop.}\ |\ \mathrm{CL})$) and \gls{tc} ($\Entr(\mathrm{isop.}\ |\ \mathrm{TC})$) information. Vertical dashed lines denote the mean value $\mu$ of the related distribution.}
    \label{fig:entropy_spatial}
\end{figure}

Further investigations into the statistical relationships between patches reveal that the entropy of isoprene emission patches conditioned to \gls{lc} data, i.e., $\Entr(\mathrm{isop.}\ |\ \mathrm{CL})$ and $\Entr(\mathrm{isop.}\ |\ \mathrm{TC})$, are higher and more narrowly distributed than that conditioned to \gls{lai}, i.e., $\Entr(\mathrm{isop.}\ |\ \mathrm{LAI})$, as illustrated in Figure~\ref{fig:entropy_spatial}. 
Among \gls{lc} types, \gls{tc} information leads to the highest average ($\mu = 3.68$), while \gls{cl} data have a slightly higher median value ($M = 3.83$) compared to \gls{tc} ($M = 3.79$). Assuming that higher conditional entropy indicates more significant potential for improving \gls{sr} learning, these findings suggest that both \gls{tc} and \gls{cl} data could contribute to enhanced \gls{sr} performance. 
%In the next section, we show how incorporating these priors helps the model to better understand the underlying \gls{lc} dynamics, leading to more precise predictions and improved estimation performance.

% Overall, these results indicate that incorporating land cover data may enrich the model, making cropland and tree cover data promising candidates for fusion with isoprene emissions to enhance \gls{sr} performance.

\subsection{Leveraging Semantic Priors}
\label{ssec:leveraging_semantic_priors}
In this section, we evaluate the effectiveness of our proposed methodology in integrating semantic information from \gls{lc} data as emission drivers to enhance the spatial resolution of isoprene emission maps. 

For a more detailed analysis, we assess the performance of our proposed methodology by training a separate system for each of the four class-specific dataset folds $\Foldn$. This results in four distinct training instances, each excluding patches from a specific climate class while maintaining the same system architecture (see Section~\ref{sssec:standard_scenario} for more details).
We start evaluating results by 
%Then, the evaluation is performed 
using patches from the test partition of the standard scenario reported in Section~\ref{sssec:standard_scenario}.
%In this way, we aim to capture better the spatial variability driven by environmental factors such as temperature, sunlight, and vegetation while enabling region-specific insights into model performance.

%Then, in Section~\ref{ssec:generalization_studies}, we assess the model's ability to super-resolve emissions not encountered during training. Specifically, we evaluate its performance on (i) emissions from unseen geographical areas and (ii) emissions from unseen climate zones.

% $\S$, i.e., the one without considering the hold-out strategies. Therefore, the scenario uses emission patches with a geographical and climate overlap with the ones used for training the \gls{sr} model over the entire study area.

We use our previous approach proposed in~\cite{giganti2023icip} as a baseline. In this approach, the \gls{hr} isoprene emission map is estimated solely from its corresponding \gls{lr} version, without incorporating any additional emission drivers to guide the \gls{sr} process. Therefore the super-resolved isoprene map is computed as
\begin{equation}
\label{eq:icip}
    \HRhat = \Tinv(\Net(\LRt)).
\end{equation}

\textbf{Transformed data domain.} To simplify, we start performing an analysis in the transformed data domain,
%(denoted as $\Tdatadom$), %(i.e., $\HRt \in \Tdatadom$), 
considering emissions after the application of the data transformation, i.e., directly comparing the \gls{hr} transformed emission $\HRt = \T(\HR)$ and the output of the \gls{sr} network $\HRhatt$. 
%For this study, the \gls{nmse} is employed to assess the model performance across different configurations as it proved to be the most sensitive among all the considered metrics (see Section~\ref{ssec:metrics}).
We compare the performance by computing the \gls{nir}, defined as
\begin{equation}
\begin{split}
    &\text{NIR}(\HRt, \HRhatt, \HRhatRef) = \\
    &\quad \frac{\text{NMSE}(\HRhatt, \HRt) - \text{NMSE}(\HRhatRef, \HRt)}{\text{NMSE}(\HRhatRef, \HRt)},
\end{split}
\label{eq:nir}
\end{equation}
where $\HRhatRef = \Net( \LRt )$ is the super-resolved isoprene emission estimated by our baseline method~\cite{giganti2023icip}, thus using only the \gls{lr} isoprene emission $\LR$ as input
%, as reported in~\eqref{eq:icip}~\cite{giganti2023icip}, 
without considering any additional emission drivers $\DLR$. 
Instead, $\HRhatt$ is the super-resolved emission obtained using our proposed methodology, exploiting information from additional \gls{lc} data.
%, and $\HRt = \T(\HR)$ is the original emission after the application of the data transformation~\cite{giganti2023icip}. 
%This metric represents the relative improvement in performance compared to a reference.

\begin{figure}
    \centering
    \includegraphics[width=0.90\linewidth]{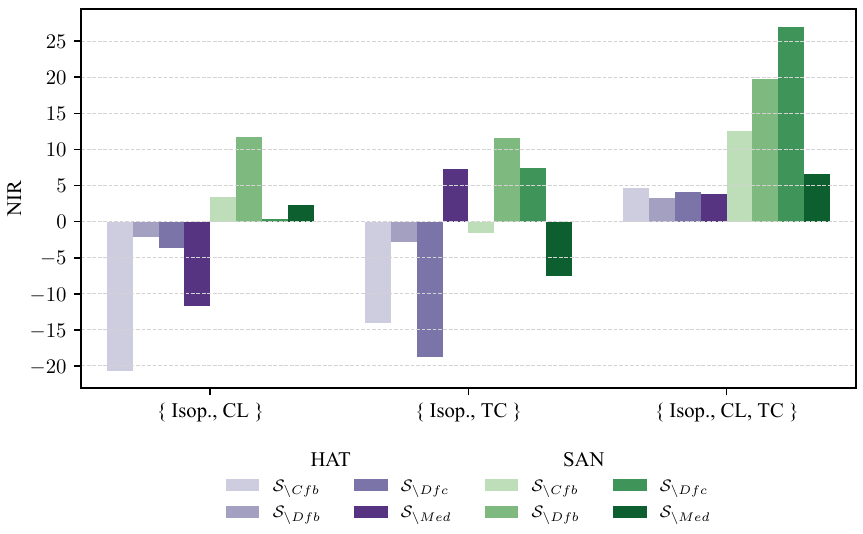}
    \caption{\gls{nir} from both the \gls{sr} networks, considering the tree proposed input configurations and the four folds $\Foldn$. The improvement is computed using~\eqref{eq:nir}, considering the $\Confone$ configuration (1-channel) in~\eqref{eq:method} as a reference method.}
    \label{fig:perc_improvement_T}
\end{figure}

In Figure~\ref{fig:perc_improvement_T}, we report the result of this study for both the considered \gls{sr} networks, i.e., \gls{san} and \gls{hat}.

For the \gls{san}, the 3-channel configuration (i.e., $\Confthree$) consistently outperforms the other setups across all folds. 
The 2-channel experiments with \gls{cl} data ($\Conftwocl$) shows improved performance. In contrast, the 2-channel experiments with \gls{tc} data ($\Conftwotc$) vary significantly, showing benefits in two of the four folds but lacking consistency overall. 

The \gls{hat} model shows a different pattern; 2-channel configurations generally decrease performance, while 3-channel configurations show stable and consistent improvements across folds. The most varying case is encountered in the $\Climfour$ fold, i.e., $\Foldfour$, where the 2-channel \gls{cl} and \gls{tc} cases yield opposite trends for both \gls{san} and \gls{hat} models, but the 3-channel cases both improve their performance. 
%Interestingly, this exception (\gls{tc} data in 2-channel configuration, $\Conftwotc$) produced the only instance of improved performance, contrasting with \gls{san}, where this configuration performed poorly. Also, the 2-channel \gls{hat} performances tended to behave in the opposite manner relative to the other folds, compared to \gls{san} 2-channel experiments.

These findings reflect that performance characteristics of different \gls{dl} architectures may vary significantly on the same data. Specifically, the negatively correlated \gls{cl} data benefits the \gls{san} model. In contrast, the same data detracts from \gls{hat}'s performance concerning the single channel reference. 
Overall, the 3-channel experiments benefit in all cases from including the additional structural information shown in the \gls{lc} maps.

\textbf{Isoprene data domain. }We now perform an analysis on the isoprene data domain, 
%which we denote with $\Idatadom$ (i.e., $\HR \in \Idatadom$), 
thus considering emissions after the back transformation of the \gls{sr} output in the isoprene data range, i.e., $\HRhat = \Tinv(\HRhatt)$. 

\begin{figure*}[ht]
    \centering
    \begin{subfigure}{1\columnwidth}
        \centering
        \includegraphics[width=\columnwidth]{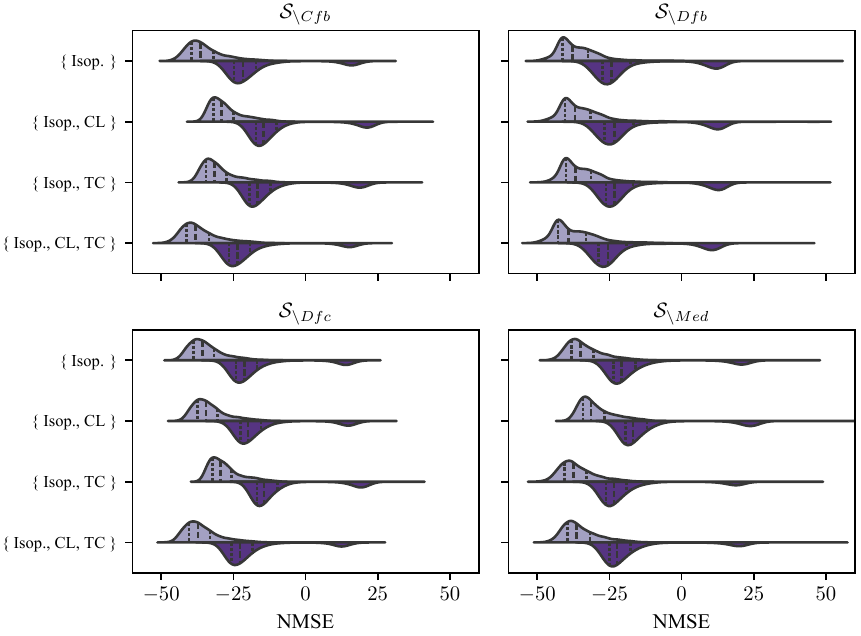}
        \caption{\gls{hat}}
        \label{fig:pre_post_hat}
    \end{subfigure}
    \hfill
    \begin{subfigure}{1\columnwidth}
        \centering
        \includegraphics[width=0.86\columnwidth]{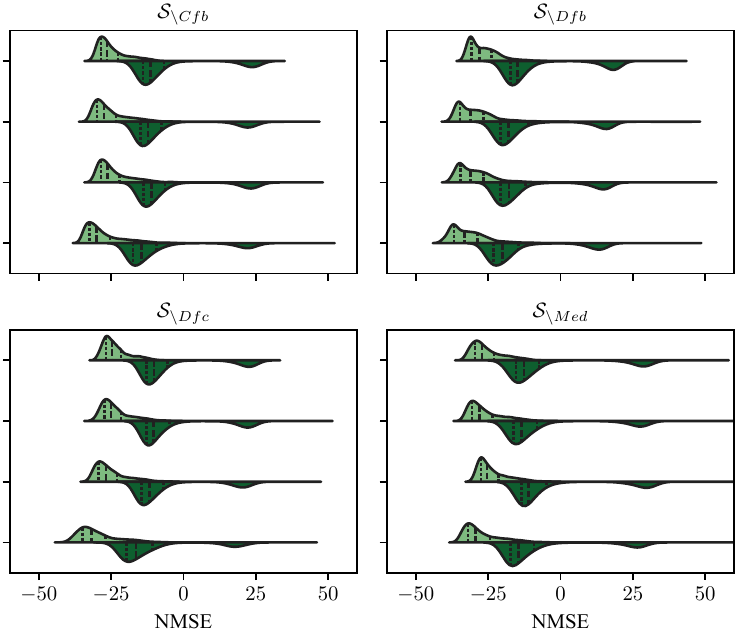}
        \caption{\gls{san}}
        \label{fig:pre_post_san}
    \end{subfigure}
    \caption{Violin plot for the transformed (upper) and isoprene (bottom) emission data domain, for the \gls{hat} (\ref{fig:pre_post_hat}) and \gls{san} (\ref{fig:pre_post_san}) network, considering the four different input configurations and folds $\Foldn$.}
    \label{fig:distribution_I_and_T}
\end{figure*}

In Figure~\ref{fig:distribution_I_and_T}, we report the \gls{nmse} considering isoprene emissions for each fold. 
%We decided to show the \gls{medfld} since it encompasses various climate classes in our study area and is, therefore, a good indicator for the overall performance of our proposed methodology.
Comparing the results before (upper distribution, using $\HRhatt$) and after applying the inverse data transformation (bottom distribution, using $\HRhat$),
%applying the inverse data transformation $\Tinv(\cdot)$, 
we can notice a marked drop in performance. 
This drop occurs for both the networks, in all folds and across all configurations. We believe it is due to the non-linear mapping introduced by the data transformation $\T(\cdot)$ and its inverse $\Tinv(\cdot)$.
Although the improvements concerning the isoprene-only scenario (1-channel, $\Confone$) remain consistent in their trends, a performance deterioration can be observed for both the \gls{sr} networks if compared to the transformed domain.

\begin{figure*}[ht]
    \centering
    \includegraphics[width=0.99\linewidth]{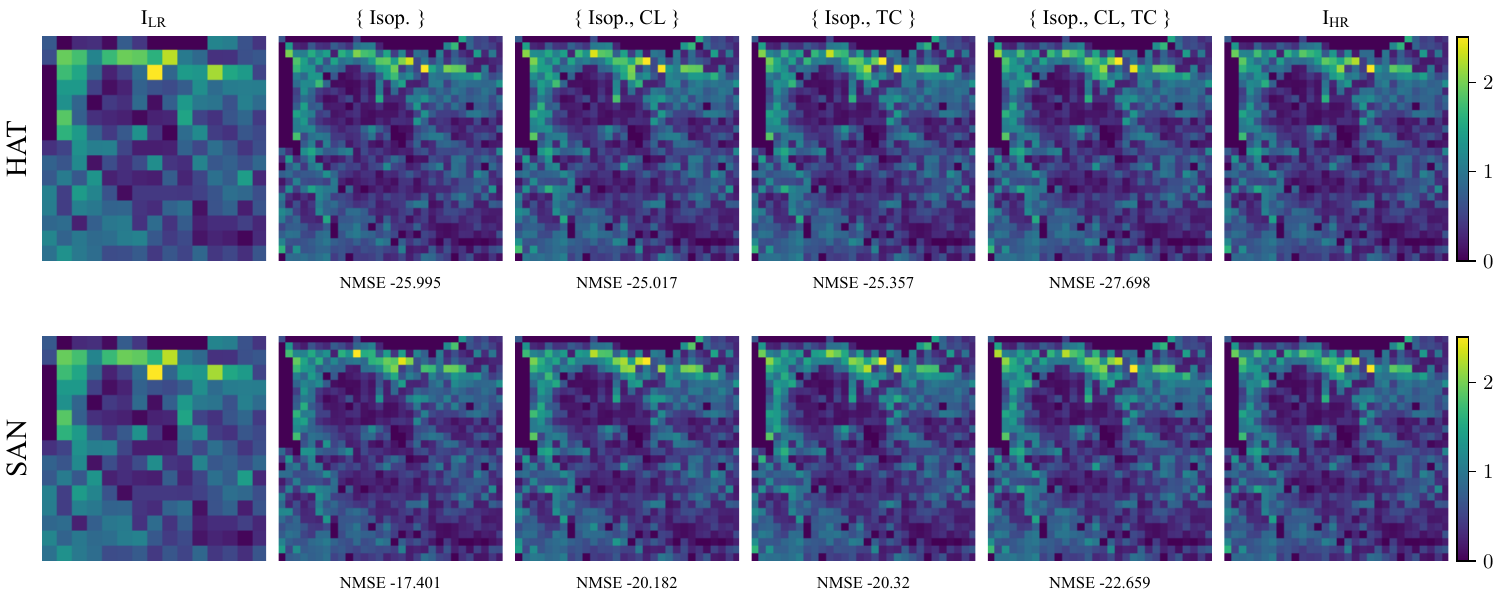}
    \caption{Super-resolved emission map examples ($\HRhat$) from both the considered \gls{sr} networks and all the four different input configurations. The first column shows the \gls{lr} emission $\LR$t; the last column report the ground-truth emission $\HR$. The other columns report $\HRhat$ for each configurations. Emissions are reported as $\frac{mol}{cm^{2}s{1}}$.}
    \label{fig:sr_examples}
\end{figure*}

In Figure~\ref{fig:sr_examples}, we report a super-resolved emission example $\HRhat$ taken from the $\Foldone$ fold from both the \gls{sr} networks and considering all the input configurations. In addition to the super-resolved emission $\HRhat$, we include its related $\LR$ (first column, input) and $\HR$ (last column, ground-truth) emissions. Notice that the three-channel configuration reports the best reconstruction results, in terms of \gls{nmse}.

\textbf{Complete metrics analysis. }For a complete analysis of all the metrics, we report the \gls{nmse}, \gls{ssim}, \gls{psnr}, \gls{maxae}, \gls{uiqi}, and \gls{scc} for all the considered configurations, before ($\HRhatt$) and after ($\HRhat$) the application of the inverse data transformation in Table~\ref{tab:T_domain_results} and Table~\ref{tab:I_domain_results}, respectively.

We can observe that \gls{hat} significantly outperforms \gls{san}. Additionally, except for $\Foldfour$, the $\Confthree$ configuration performs best, while the second-best is the baseline configuration, i.e., $\Confone$. However, for $\Foldfour$, the best-performing configuration is the 2-channel $\Conftwotc$, followed by the 3-channel configuration, i.e., $\Confthree$.

Across all input configurations, \gls{hat} demonstrates more stable performance than \gls{san}, as evidenced by lower standard deviations in metrics like \gls{ssim} and higher consistency in \gls{maxae} and \gls{uiqi} across data folds. This stability can be attributed to their distinct attention mechanisms. \gls{san}, which leverages second-order channel attention, is more sensitive to high-frequency components. While this enhances fine details, it also amplifies noise when handling unseen data.

These results are consistent both before (Table~\ref{tab:T_domain_results}, using $\HRhatt$) and after (Table~\ref{tab:I_domain_results}, using $\HRhat$) applying the inverse data transformation.

\begin{table*}[ht]
    \caption{\gls{sr} performance results evaluated on $\HRhatt$, i.e., transformed data domain. \textit{Avg} and \textit{Std} denote the average and standard deviation of the specific metric, respectively. In \textbf{bold}, we denote the best-performing average value, and \underline{underlined} values denote the second-best, for each fold $\Foldn$ for both the \gls{sr} networks. Arrow direction ($\uparrow$ or $\downarrow$) indicates the preferred direction for each metric.}
    \centering
    %\resizebox{1.0\textwidth}{!}{
    \begin{tabular}{lllcccccc}
        \toprule
        \multicolumn{9}{c}{\textbf{Transformed Data Domain (\textit{Avg}, \textit{Std})}} \\
        \midrule
        \textbf{Model} & \textbf{Data Fold} & \textbf{Configuration} & \textbf{NMSE $\downarrow$}  & \textbf{SSIM $\uparrow$}  & \textbf{PSNR $\uparrow$}  & \textbf{MaxAE $\downarrow$}  & \textbf{UIQI $\uparrow$}  & \textbf{SCC $\uparrow$}  \\
        \toprule
        \textbf{HAT} & $\Foldone$ & $\Confone$ & \underline{-35.04}, 6.42 & \underline{0.98}, 0.05 & \underline{38.11}, 5.02 & \underline{0.04}, 0.04 & \underline{0.97}, 0.09 & \underline{0.98}, 0.06 \\
         &  & $\Conftwocl$ & -27.79, 5.71 & 0.94, 0.08 & 30.86, 4.29 & 0.10, 0.06 & 0.94, 0.11 & 0.94, 0.09 \\
         &  & $\Conftwotc$ & -30.14, 6.02 & 0.96, 0.07 & 33.21, 4.61 & 0.07, 0.05 & 0.96, 0.10 & 0.96, 0.07 \\
         &  & $\Confthree$ & \textbf{-36.69}, 6.48 & \textbf{0.99}, 0.04 & \textbf{39.76}, 5.05 & \textbf{0.04}, 0.04 & \textbf{0.98}, 0.09 & \textbf{0.99}, 0.04 \\
         \cmidrule(lr){2-9}
         & $\Foldtwo$ & $\Confone$ & \underline{-36.23}, 7.04 & \underline{0.98}, 0.06 & \underline{39.69}, 5.55 & \underline{0.03}, 0.02 & \underline{0.98}, 0.10 & \underline{0.98}, 0.06 \\
         &  & $\Conftwocl$ & -35.43, 6.93 & 0.98, 0.07 & 38.89, 5.37 & 0.04, 0.03 & 0.97, 0.11 & 0.98, 0.07 \\
         &  & $\Conftwotc$ & -35.21, 6.63 & 0.98, 0.06 & 38.67, 5.11 & 0.04, 0.03 & 0.97, 0.10 & 0.98, 0.06 \\
         &  & $\Confthree$ & \textbf{-37.42}, 7.22 & \textbf{0.99}, 0.05 & \textbf{40.89}, 5.58 & \textbf{0.03}, 0.02 & \textbf{0.98}, 0.10 & \textbf{0.99}, 0.05 \\
         \cmidrule(lr){2-9}
         & $\Foldthree$ & $\Confone$ & \underline{-34.68}, 5.80 & \underline{0.99}, 0.04 & \underline{37.88}, 4.45 & \underline{0.04}, 0.04 & \underline{0.98}, 0.08 & \underline{0.98}, 0.05 \\
         &  & $\Conftwocl$ & -33.39, 5.79 & 0.98, 0.04 & 36.6, 4.43 & 0.05, 0.04 & 0.98, 0.08 & 0.98, 0.05 \\
         &  & $\Conftwotc$ & -28.16, 5.42 & 0.95, 0.07 & 31.36, 4.03 & 0.08, 0.06 & 0.95, 0.09 & 0.95, 0.07 \\
         &  & $\Confthree$ & \textbf{-36.12}, 5.89 & \textbf{0.99}, 0.03 & \textbf{39.32}, 4.55 & \textbf{0.04}, 0.03 & \textbf{0.98}, 0.08 & \textbf{0.99}, 0.04 \\
         \cmidrule(lr){2-9}
         & $\Foldfour$ & $\Confone$ & -33.81, 6.17 & 0.98, 0.04 & 36.84, 4.82 & 0.05, 0.04 & 0.97, 0.08 & 0.98, 0.05 \\
         &  & $\Conftwocl$ & -29.85, 5.92 & 0.96, 0.06 & 32.89, 4.55 & 0.07, 0.05 & 0.96, 0.09 & 0.96, 0.07 \\
         &  & $\Conftwotc$ & \textbf{-36.28}, 6.13 & \textbf{0.99}, 0.03 & \textbf{39.31}, 4.80 & \textbf{0.04}, 0.03 & \textbf{0.98}, 0.08 & \textbf{0.99}, 0.03 \\
         &  & $\Confthree$ & \underline{-35.10}, 6.22 & \underline{0.99}, 0.04 & \underline{38.14}, 4.88 & \underline{0.04}, 0.04 & \underline{0.98}, 0.08 & \underline{0.98}, 0.04 \\
         \toprule
        \textbf{SAN} & $\Foldone$ & $\Confone$ & -24.93, 4.90 & 0.90, 0.11 & 28.00, 3.67 & 0.13, 0.09 & 0.89, 0.13 & 0.88, 0.13 \\
         &  & $\Conftwocl$ & -25.80, 5.56 & 0.91, 0.11 & 28.87, 4.30 & 0.12, 0.08 & 0.91, 0.12 & 0.90, 0.11 \\
         &  & $\Conftwotc$ & -24.52, 5.42 & 0.90, 0.11 & 27.59, 4.12 & 0.14, 0.10 & 0.90, 0.13 & 0.89, 0.11 \\
         &  & $\Confthree$ & -28.07, 6.17 & 0.94, 0.09 & 31.14, 4.85 & 0.09, 0.06 & 0.94, 0.11 & 0.94, 0.09 \\
         \cmidrule(lr){2-9}
         & $\Foldtwo$ & $\Confone$ & -26.84, 5.00 & 0.94, 0.10 & 30.30, 3.73 & 0.10, 0.08 & 0.94, 0.12 & 0.93, 0.12 \\
         &  & $\Conftwocl$ & -29.99, 5.89 & 0.96, 0.09 & 33.45, 4.62 & 0.08, 0.07 & 0.96, 0.11 & 0.96, 0.09 \\
         &  & $\Conftwotc$ & -29.96, 5.88 & 0.96, 0.09 & 33.42, 4.55 & 0.07, 0.07 & 0.96, 0.11 & 0.96, 0.08 \\
         &  & $\Confthree$ & -32.13, 6.00 & 0.97, 0.07 & 35.59, 4.60 & 0.06, 0.06 & 0.97, 0.1 & 0.97, 0.07 \\
         \cmidrule(lr){2-9}
         & $\Foldthree$ & $\Confone$ & -23.62, 4.39 & 0.89, 0.10 & 26.83, 3.18 & 0.17, 0.12 & 0.89, 0.13 & 0.87, 0.13 \\
         &  & $\Conftwocl$ & -23.71, 5.04 & 0.89, 0.11 & 26.92, 3.82 & 0.18, 0.12 & 0.90, 0.12 & 0.89, 0.11 \\
         &  & $\Conftwotc$ & -25.37, 5.42 & 0.92, 0.10 & 28.57, 4.14 & 0.13, 0.09 & 0.92, 0.11 & 0.91, 0.1 \\
         &  & $\Confthree$ & -30.01, 6.67 & 0.96, 0.08 & 33.21, 5.35 & 0.07, 0.06 & 0.95, 0.1 & 0.96, 0.07 \\
         \cmidrule(lr){2-9}
         & $\Foldfour$ & $\Confone$ & -25.76, 5.23 & 0.91, 0.10 & 28.79, 4.05 & 0.13, 0.09 & 0.91, 0.13 & 0.90, 0.13 \\
         &  & $\Conftwocl$ & -26.33, 5.71 & 0.92, 0.10 & 29.37, 4.51 & 0.12, 0.09 & 0.92, 0.12 & 0.92, 0.11 \\
         &  & $\Conftwotc$ & -23.81, 5.09 & 0.89, 0.11 & 26.84, 3.90 & 0.17, 0.11 & 0.90, 0.13 & 0.88, 0.11 \\
         &  & $\Confthree$ & -27.44, 6.18 & 0.94, 0.10 & 30.48, 4.93 & 0.10, 0.07 & 0.93, 0.11 & 0.93, 0.09 \\
        \bottomrule
    \end{tabular}
    %}
    \label{tab:T_domain_results}
\end{table*}

\begin{table*}[ht]
    \caption{\gls{sr} performance results evaluated on $\HRhat$, i.e., isoprene data domain. \textit{Avg} and \textit{Std} denote the average and standard deviation of the specific metric, respectively. In \textbf{bold}, we denote the best-performing average value, and \underline{underlined} values denote the second-best, for each fold $\Foldn$ for both the \gls{sr} networks. Arrow direction ($\uparrow$ or $\downarrow$) indicates the preferred direction for each metric.}
    \centering
    %\resizebox{1.0\textwidth}{!}{
    \begin{tabular}{lllcccccc}
        \toprule
        \multicolumn{9}{c}{\textbf{Isoprene Data Domain (\textit{Avg}, \textit{Std})}} \\
        \midrule
        \textbf{Model} & \textbf{Data Fold} & \textbf{Configuration} & \textbf{NMSE $\downarrow$}  & \textbf{SSIM $\uparrow$}  & \textbf{PSNR $\uparrow$}  & \textbf{MaxAE $\downarrow$}  & \textbf{UIQI $\uparrow$}  & \textbf{SCC $\uparrow$}  \\
        \toprule
        \textbf{HAT} & $\Foldone$ & $\Confone$ & \underline{-17.35}, 12.83 & \underline{0.94}, 0.15 & \underline{29.85}, 11.93 & \underline{204.68}, 555.71 & \underline{0.70}, 0.40 & \underline{0.94}, 0.11 \\
         &  & $\Conftwocl$ & -9.17, 13.51 & 0.85, 0.21 & 21.67, 12.83 & 281.33, 643.93 & 0.62, 0.40 & 0.86, 0.14 \\
         &  & $\Conftwotc$ & -11.83, 12.96 & 0.89, 0.19 & 24.33, 12.22 & 250.26, 610.56 & 0.66, 0.40 & 0.90, 0.12 \\
         &  & $\Confthree$ & \textbf{-19.42}, 12.58 & \textbf{0.95}, 0.14 & \textbf{31.92}, 11.76 & \textbf{182.35}, 526.54 & \textbf{0.71}, 0.40 & \textbf{0.96}, 0.10 \\
         \cmidrule(lr){2-9}
         & $\Foldtwo$ & $\Confone$ & \underline{-17.80}, 15.17 & \underline{0.92}, 0.18 & \underline{30.03}, 14.40 & 345.73, 706.11 & \underline{0.84}, 0.29 & \underline{0.94}, 0.13 \\
         &  & $\Conftwocl$ & -17.07, 14.99 & 0.91, 0.18 & 29.30, 14.19 & \underline{345.61}, 706.91 & 0.83, 0.29 & 0.93, 0.14 \\
         &  & $\Conftwotc$ & -16.80, 14.85 & 0.91, 0.18 & 29.03, 14.15 & 346.54, 707.77 & 0.83, 0.30 & 0.93, 0.13 \\
         &  & $\Confthree$ & \textbf{-19.47}, 14.63 & \textbf{0.93}, 0.15 & \textbf{31.70}, 13.87 & \textbf{318.25}, 679.18 & \textbf{0.85}, 0.28 & \textbf{0.94}, 0.13 \\
         \cmidrule(lr){2-9}
         & $\Foldthree$ & $\Confone$ & \underline{-16.93}, 12.30 & \underline{0.94}, 0.15 & \underline{29.82}, 11.58 & \underline{149.16}, 386.15 & \underline{0.70}, 0.40 & \underline{0.95}, 0.10 \\
         &  & $\Conftwocl$ & -15.47, 12.25 & 0.93, 0.17 & 28.36, 11.59 & 154.12, 392.45 & 0.69, 0.40 & 0.94, 0.10 \\
         &  & $\Conftwotc$ & -9.04, 12.86 & 0.86, 0.22 & 21.93, 12.36 & 199.73, 442.25 & 0.63, 0.40 & 0.88, 0.13 \\
         &  & $\Confthree$ & \textbf{-18.76}, 11.82 & \textbf{0.95}, 0.14 & \textbf{31.65}, 11.16 & \textbf{131.37}, 363.45 & \textbf{0.71}, 0.39 & \textbf{0.96}, 0.09 \\
         \cmidrule(lr){2-9}
         & $\Foldfour$ & $\Confone$ & -15.72, 14.09 & 0.93, 0.17 & 27.54, 13.31 & 217.76, 577.69 & 0.67, 0.41 & 0.93, 0.12 \\
         &  & $\Conftwocl$ & -10.95, 14.75 & 0.88, 0.21 & 22.77, 14.09 & 263.47, 629.47 & 0.63, 0.41 & 0.89, 0.14 \\
         &  & $\Conftwotc$ & \textbf{-18.98}, 13.35 & \textbf{0.95}, 0.14 & \textbf{30.79}, 12.57 & \textbf{182.40}, 531.77 & \textbf{0.69}, 0.40 & \textbf{0.96}, 0.10 \\
         &  & $\Confthree$ & \underline{-17.52}, 13.71 & \underline{0.94}, 0.16 & \underline{29.33}, 12.97 & \underline{196.47}, 550.80 & \underline{0.68}, 0.41 & \underline{0.95}, 0.11 \\
         \toprule
        \textbf{SAN} & $\Foldone$ & $\Confone$ & -5.34, 14.03 & 0.77, 0.24 & 17.84, 13.54 & 337.51, 695.52 & 0.55, 0.39 & 0.76, 0.19 \\
         &  & $\Conftwocl$ & -6.47, 13.61 & 0.80, 0.23 & 18.97, 13.31 & 314.10, 675.29 & 0.59, 0.39 & 0.79, 0.18 \\
         &  & $\Conftwotc$ & -5.17, 13.77 & 0.77, 0.24 & 17.67, 13.34 & 326.66, 686.39 & 0.56, 0.39 & 0.76, 0.17 \\
         &  & $\Confthree$ & -8.86, 13.96 & 0.84, 0.24 & 21.37, 13.52 & 281.56, 644.70 & 0.62, 0.40 & 0.85, 0.15 \\
         \cmidrule(lr){2-9}
         & $\Foldtwo$ & $\Confone$ & -7.54, 14.41 & 0.80, 0.26 & 19.77, 13.95 & 450.83, 793.87 & 0.72, 0.34 & 0.81, 0.18 \\
         &  & $\Conftwocl$ & -11.54, 14.07 & 0.86, 0.22 & 23.77, 13.65 & 380.77, 739.76 & 0.78, 0.31 & 0.88, 0.16 \\
         &  & $\Conftwotc$ & -11.90, 13.49 & 0.87, 0.21 & 24.13, 12.94 & 364.04, 726.35 & 0.79, 0.31 & 0.88, 0.15 \\
         &  & $\Confthree$ & -14.21, 13.66 & 0.89, 0.19 & 26.44, 13.09 & 345.90, 708.96 & 0.81, 0.30 & 0.91, 0.14 \\
         \cmidrule(lr){2-9}
         & $\Foldthree$ & $\Confone$ & -4.18, 13.46 & 0.76, 0.25 & 17.07, 13.06 & 248.90, 485.88 & 0.53, 0.39 & 0.75, 0.18 \\
         &  & $\Conftwocl$ & -4.05, 13.41 & 0.76, 0.25 & 16.94, 13.16 & 244.59, 481.94 & 0.56, 0.39 & 0.77, 0.17 \\
         &  & $\Conftwotc$ & -6.25, 12.95 & 0.80, 0.24 & 19.14, 12.52 & 220.32, 462.37 & 0.59, 0.40 & 0.81, 0.16 \\
         &  & $\Confthree$ & -11.44, 13.15 & 0.88, 0.22 & 24.32, 12.58 & 179.76, 423.17 & 0.66, 0.40 & 0.90, 0.13 \\
         \cmidrule(lr){2-9}
         & $\Foldfour$ & $\Confone$ & -5.40, 16.36 & 0.79, 0.26 & 17.21, 15.94 & 339.18, 699.78 & 0.55, 0.40 & 0.79, 0.19 \\
         &  & $\Conftwocl$ & -6.68, 15.72 & 0.81, 0.26 & 18.50, 15.35 & 306.36, 671.01 & 0.57, 0.41 & 0.82, 0.17 \\
         &  & $\Conftwotc$ & -3.21, 16.32 & 0.75, 0.27 & 15.02, 16.09 & 354.69, 711.83 & 0.53, 0.40 & 0.77, 0.17 \\
         &  & $\Confthree$ & -8.18, 15.28 & 0.83, 0.25 & 19.99, 14.86 & 279.92, 646.75 & 0.60, 0.41 & 0.85, 0.16 \\
        \bottomrule
    \end{tabular}
    %}
    \label{tab:I_domain_results}
\end{table*}

Moreover, in Table~\ref{tab:runtime}, we report the training times and epochs required for convergence for each experiment. It is worth noting that adding additional information, such as emission drivers in $\DLR$, often increases the complexity of the model. Although additional inputs can provide physically relevant constraints that, in theory, help regularize the learning process and reduce the inherent ill-posedness of the \gls{sr} problem, this benefit is not always observed in practice. Indeed, the increased complexity typically leads to longer runtimes and more epochs needed to reach convergence. However, the 2-channel configurations, i.e., $\Conftwocl$ and $\Conftwotc$, are generally faster than both the 1-channel, i.e., $\Confone$, and 3-channel, i.e., $\Confthree$ configurations for both \gls{sr} networks, likely due to a better balance between information richness and computational overhead.

% Additionally, due to its higher computational complexity, the transformer-based \gls{hat} network requires more epochs and, therefore, longer training times than the \gls{san} network. The convergence behavior also appears highly dependent on the data fold used for training. Since each data fold represents patches of isoprene emissions from different climate zones, the variability in emission patterns and environmental conditions across these zones likely influences the learning dynamics. For instance, folds corresponding to regions with more complex or heterogeneous emission profiles may require more epochs to achieve convergence. In contrast, folds from more uniform or predictable regions may converge faster. This variability highlights the need for geographical and climatic diversity in training data when designing and evaluating \gls{sr} models.
% Despite the regularization potential of additional inputs, the lack of a consistent reduction in convergence time can be attributed to the increased model complexity and the need to learn more intricate patterns from the additional data. 
% While the additional inputs provide valuable physical constraints, they also introduce more parameters and interactions, underscoring the importance of carefully balancing the amount of input information with the computational cost and convergence efficiency.

The transformer-based \gls{hat} network, due to its higher computational complexity, requires more epochs and longer training times than the \gls{san} network. Its convergence depends on the data fold used, as different climate zones exhibit varying emission patterns. Regions with complex emissions may need more epochs, while uniform areas converge faster. This highlights the importance of diverse training data. While additional inputs offer physical constraints, they also increase model complexity, requiring a careful balance between input information, computational cost, and convergence efficiency.

\begin{table}[ht]
    \caption{Training runtimes of all the experiments, therefore considering all the different configurations, folds, and \gls{sr} networks.}
    \centering
    \resizebox{0.9\columnwidth}{!}{
    \begin{tabular}{llccccc}
        \toprule
        \multicolumn{7}{c}{\textbf{Training Runtimes}} \\
        \midrule
        & & \multicolumn{2}{c}{\textbf{HAT}} & & \multicolumn{2}{c}{\textbf{SAN}} \\
        \midrule
        \textbf{Data Fold} & \textbf{Configuration} & \textbf{Epochs} & \textbf{Time [min]} & & \textbf{Epochs} & \textbf{Time [min]}  \\
        \toprule
        $\Foldone$ & $\Confone$ & 499 & 1847 & & 72 & 452 \\
        & $\Conftwocl$ & 75 & 280 & & 69 & 452 \\
        & $\Conftwotc$ & 88 & 322 & & 66 & 430 \\
        & $\Confthree$ & 307 & 1148 & & 72 & 490 \\
        \midrule
        $\Foldtwo$ & $\Confone$ & 679 & 2561 & & 67 & 440 \\
        & $\Conftwocl$ & 729 & 2828 & & 69 & 466 \\
        & $\Conftwotc$ & 666 & 2634 & & 69 & 467 \\
        & $\Confthree$ & 614 & 2290 & & 78 & 500 \\
        \midrule
        $\Foldthree$ & $\Confone$ & 499 & 1843 & & 69 & 436 \\
        & $\Conftwocl$ & 499 & 1846 & & 64 & 439 \\
        & $\Conftwotc$ & 77 & 280 & & 67 & 442 \\
        & $\Confthree$ & 499 & 1840 & & 149 & 1043 \\
        \midrule
        $\Foldfour$ & $\Confone$ & 499 & 1859 & & 72 & 487 \\
        & $\Conftwocl$ & 85 & 311 & & 69 & 466 \\
        & $\Conftwotc$ & 483 & 1798 & & 65 & 430 \\
        & $\Confthree$ & 499 & 1854 & & 69 & 438 \\
        \bottomrule
    \end{tabular}
    }
    \label{tab:runtime}
\end{table}

% \subsection{Perfect Prognosis Scenario}
% \label{ssec:perfect_prognosis_scenario}
% [TODO]

\subsection{Generalization Studies}
\label{ssec:generalization_studies}
To assess the model capabilities in super-resolving unseen emissions during training, we perform generalization studies addressing the two additional scenarios presented in Section~\ref{ssec:experimental_scenarios} that are: (i) the unseen spatial areas scenario; (ii) the unseen climate zones scenario. 
We perform these investigations for each $\Foldn$ fold of the entire experimental dataset $\AllSet$.

% SPATIAL
For the unseen spatial areas scenario (Section~\ref{sssec:spatial_scenario}), isoprene emissions from a specific geographical region with the same climate classes as the training data 
are used in the test phase.
This is done to assess the model's ability to generalize to different geographical areas while keeping climate conditions consistent with those seen during training. 

%We perform this study for each $\Foldn$ fold of the complete experimental dataset, thus addressing four different unseen spatial areas.

% CLIMATE
Regarding the unseen climate zones scenario (Section~\ref{sssec:climate_scenario}) instead, the test emissions refer exclusively to a specific climate class not used in training, forcing the model to super-resolve emissions from an unseen climate zone. 
%The climate scenario is expected to pose a more significant challenge than the spatial one due to the unseen structural patterns in the input emission and the inherent differences in the underlying isoprene emission mechanisms, with different species contributing emissions with different spatial patterns in diverse climate zones~\cite{guenther_megan3_2020, penuelas_bvocs_2010}.

%The comparison between these two scenarios is intended to evaluate whether our method is learning based on structural cues alone or if they can leverage the additional \gls{cl} and \gls{tc} maps, which encode additional structural information on the underlying vegetative speciation of the pixels, for better results in the spatial hold-out scenario (where these structural relationships have been observed in training).

\begin{figure}[ht]
    \centering
    \includegraphics[width=0.99\linewidth]{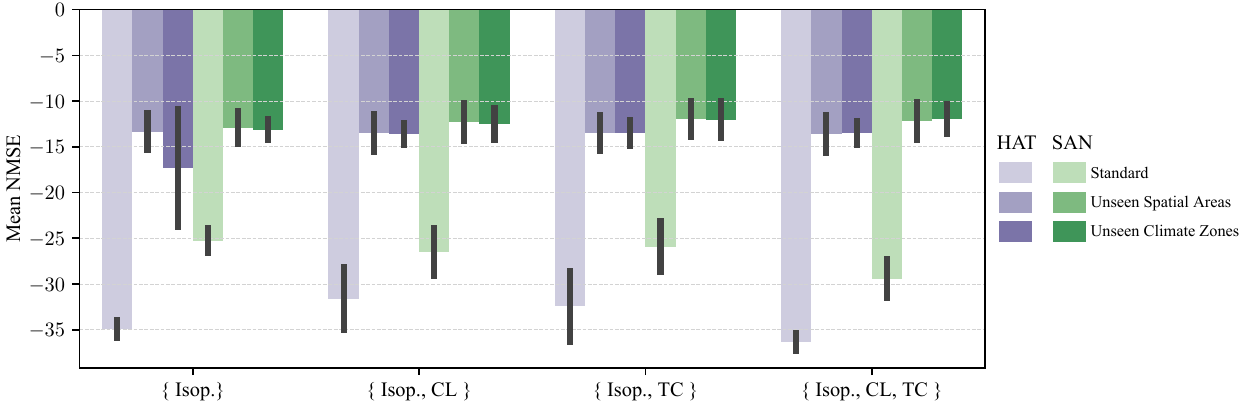}
    \caption{Results comparing the standard, unseen spatial areas and unseen climate zones scenarios, for both the \gls{sr} networks. The reported \gls{nmse} is the average \gls{nmse} among all the folds $\Foldn$ of $\AllSet$. The black segment denotes the standard deviation, for each experiment.}
    \label{fig:holdout_studies}
\end{figure}

Figure~\ref{fig:holdout_studies} reports the result obtained using the \gls{san} and the \gls{hat} networks, averaged across the four $\Foldn$ folds. We also show the test set performance when considering the standard scenario (Section~\ref{sssec:standard_scenario}).
The reported \gls{nmse} is referred to the transformed data domain,  
%\Tdatadom$
thus assessing the \gls{sr} performance on the $\HRhatt$ isoprene emissions.
%, and is averaged among all the four dataset fold $\Foldn$. 

We can notice a significant performance drop on both the unseen spatial and climate sets compared to the standard test set. 
%The average performance on these test sets decreases substantially, d
This demonstrates the models' challenges when generalizing to unseen contexts. 
%However, the results achieved \gls{nmse} are still acceptable if we compare them... 
%performance across channel configurations, thus 
Moreover, including additional priors to guide the \gls{sr} process is not helpful in both unseen sets. %The \gls{hat} model's performance remains consistent across channel configurations, while the \gls{san} model shows a minor reduction in accuracy as channels increased, though this effect is minimal. 
%Interestingly, on the standard test set, we observe that adding both \gls{tc} and \gls{cl} improves performance, as they provide complementary spatial context.
%\gls{tc} correlates strongly with emissions. 
%In contrast, \gls{cl} adds variability in unforested areas. 
%However, including just one \gls{lc} information introduces noise or redundancy without sufficient complementary information, leading to a higher standard deviation in \gls{nmse} (Figure~\ref{fig:holdout_studies}, black segments) due to inconsistent performance. Using all three features together, that is, the configuration $\Confthree$, the \gls{sr} networks effectively leverage their combined information, reducing the standard deviation of \gls{nmse}. This improvement occurs because the model can better generalize across diverse landscapes, balancing the contributions of each \gls{lc} data, and minimizing prediction uncertainty.

\begin{figure*}[ht]
    \centering    
    \begin{subfigure}{0.99\textwidth}
        \centering
        \includegraphics[width=\linewidth]{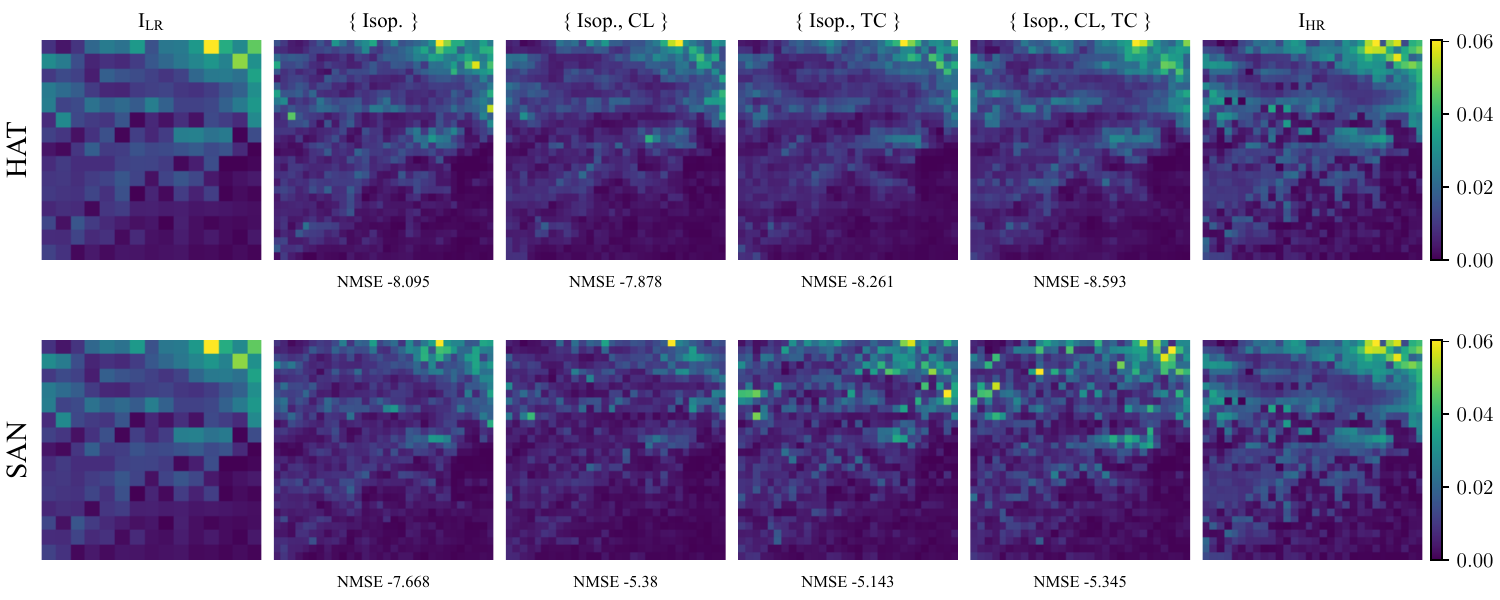}
        \caption{Unseen spatial areas scenario.}
        \label{fig:spatial_sr}
    \end{subfigure}

    \vspace{5pt} % Adjusts vertical spacing between the figures

    \begin{subfigure}{0.99\textwidth}
        \centering
        \includegraphics[width=\linewidth]{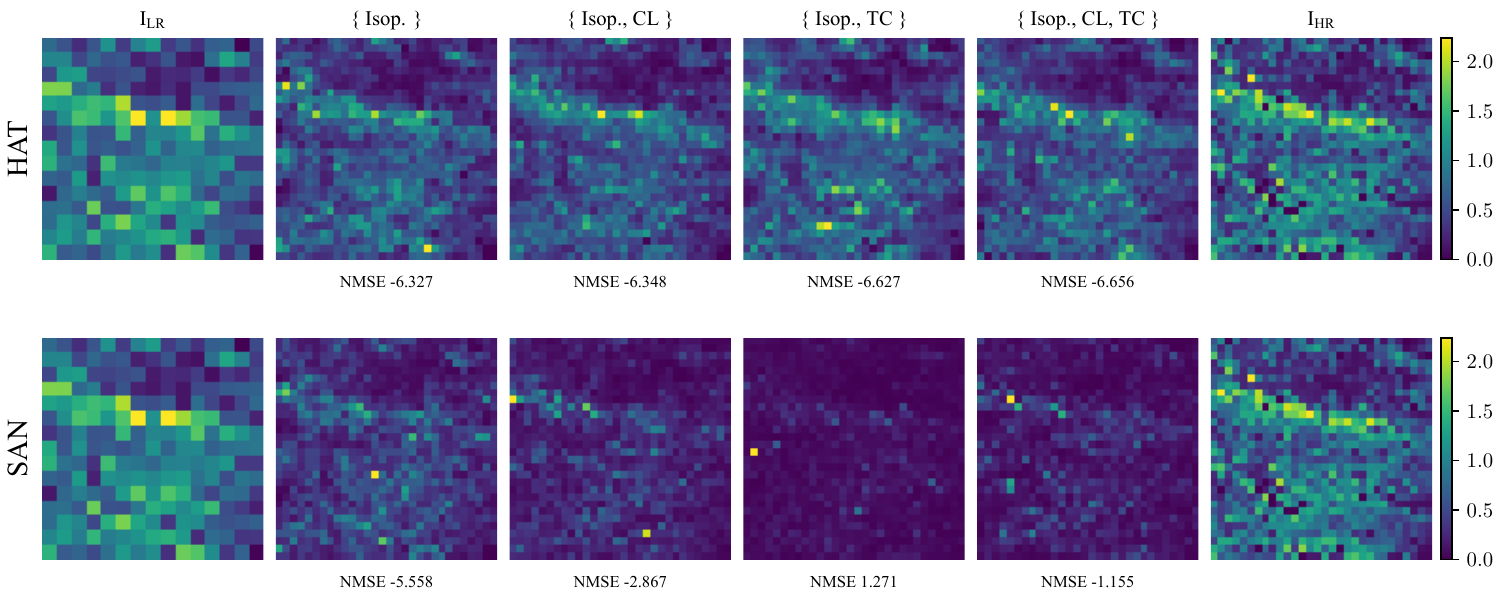}
        \caption{Unseen climate zones scenario.}
        \label{fig:climate_sr}
    \end{subfigure}
    
    \caption{Super-resolved emission map examples ($\HRhat$) from both the considered \gls{sr} networks and all the four different input configurations. Patches are related to the unseen spatial areas (\ref{fig:spatial_sr}) and climate zones (\ref{fig:climate_sr}) scenarios. The first column shows the \gls{lr} emission $\LR$; the last column reports the ground-truth emission $\HR$. Emissions are reported as $\frac{mol}{cm^{2}s{1}}$.}
    \label{fig:sr_examples_holdout_studies}
\end{figure*}

%As mentioned earlier, the average performance of both the \gls{sr} networks for the unseen spatial and climate test sets is similar. 

In Figure~\ref{fig:sr_examples_holdout_studies}, we report \gls{sr} examples to highlight the different behaviors adopted by our \gls{sr} networks when dealing with unseen emissions. 
We can notice subtle differences in reconstructing emission patterns by looking at the four super-resolved versions $\HRhat$.
Specifically, \gls{san} leads to noisier and inaccurate emission patterns in the super-resolved outputs. In contrast, \gls{hat} integrates multiple attention mechanisms, providing implicit regularization that results in smoother and more stable reconstructions. 

It is worth noticing that, even if the achieved \gls{nmse} values are significantly lower than in the standard scenario, \gls{hat} always carries more accurate \gls{sr} than \gls{san}. 
We conjecture that
%, by balancing local and global contexts, \gls{hat} reduces sensitivity to high-frequency artifacts, producing outputs that are slightly low-passed, as evident from the visual comparison between the super-resolved emissions from \gls{san} and \gls{hat} in Figure~\ref{fig:spatial_sr}. Additionally, 
the more substantial feature aggregation in \gls{hat} helps suppress noise and improves robustness when generalizing to novel patterns, as reflected in its superior performance across all configurations.

\section{Discussion}
\label{sec:discussion}
\glsreset{tc}
\glsreset{cl}
\glsreset{lc}
\glsreset{lai}

This study has shown the potential of leveraging \gls{lc} priors to enhance the spatial resolution of isoprene emission maps over Europe using \gls{dl}-based \gls{sr} techniques. By incorporating \gls{cl} and \gls{tc} data, we can augment isoprene emission predictions, providing insights into the interplay between land use characteristics and \gls{bvoc} dynamics.

% Statistical Relationships and Semantic Priors  
The statistical analysis in Section~\ref{ssec:statistical_relationships} revealed that \gls{tc} exhibited the strongest correlation with isoprene emissions, while \gls{cl} showed a moderate, albeit negative, correlation. These findings underscore the importance of vegetation type and density in determining isoprene emissions, as \gls{tc} acts as a significant source of isoprene. Interestingly, while the \gls{lai} captures seasonal variations in vegetation, its weak spatial correlation with isoprene emissions suggested it is less effective as a direct driver for spatial \gls{sr} tasks. However, the inclusion of \gls{lai} may still be beneficial for capturing temporal dynamics in future models. 

% Super-Resolution Model Performance  
The comparison of \gls{sisr} and \gls{misr} approaches highlighted the advantages of incorporating multiple drivers into the \gls{sr} process. Specifically, the \gls{hat} and \gls{san} models both exhibited improved performance when \gls{cl} and \gls{tc} data are included, as we can see from Figure~\ref{fig:perc_improvement_T}. Among the configurations tested, the 3-channel approach, i.e., including \gls{cl} and \gls{tc} information in the \gls{sr} process, consistently outperformed single-channel and 2-channel setups, demonstrating the value of integrating diverse data sources to enhance spatial predictions.  

While the 3-channel experiments did not converge in fewer epochs than the 1- and 2-channel setups, they achieved significantly higher reconstruction accuracy. This improvement can be attributed to the additional channels providing richer semantic priors, which better constrain the ill-posed nature of the \gls{sr} problem. This finding highlights the importance of leveraging diverse, complementary data sources to enhance the precision of the \gls{sr} process—even if computational efficiency remains comparable.

The observed differences in performance between the \gls{hat} and \gls{san} models can be attributed to their architectural designs. \gls{hat}, leveraging transformer-based global attention mechanisms, excels in capturing long-range dependencies and large-scale spatial patterns, making it particularly suitable for scenarios where broader geographic or environmental contexts influence emission patterns. In contrast, \gls{san}'s second-order attention mechanism is adept at modeling localized and fine-grained spatial details, providing an advantage in regions with high emission variability or complex local patterns. These complementary strengths highlight the importance of selecting architectures based on the specific characteristics of the data and the target application.  

%Generalization Challenges  
The generalization studies, which evaluated the models' performance on unseen climate zones and geographic regions, revealed a significant drop in performance compared to the standard test set. Several factors may explain this. First, the training dataset likely underrepresents certain climate zones or geographic regions, limiting the model's ability to learn robust patterns applicable to diverse contexts. Second, isoprene emissions are highly dynamic and influenced by complex environmental factors, making extrapolation to unseen conditions challenging.  

Additionally, the reliance on static \gls{lc} data may introduce temporal inconsistencies. As a matter of fact, the \gls{lc} maps represent a snapshot in time and do not account for seasonal or long-term changes in vegetation, potentially leading to misalignment with the isoprene emission dynamics in unseen regions. This temporal mismatch underscores the need for integrating dynamic priors, such as real-time vegetation indices or meteorological variables, to improve generalization performance.  

Finally, the high sparsity and variability of isoprene emissions further complicate the learning process. Sparse data, particularly in regions with low emission levels, may hinder the models' ability to generalize effectively to areas with distinct emission characteristics.  

% Implications for Atmospheric and Climate Research  
%Integrating \gls{lc} priors into \gls{sr} models offers a promising approach for improving the spatial resolution of \gls{bvoc} emission inventories. Accurate \gls{hr} isoprene maps are critical for understanding atmospheric chemistry, air quality, and climate interactions. The proposed methodology enhances the ability to model these processes, enabling more precise simulations of photochemical reactions, cloud formation, and regional climate dynamics.  

%However, the findings also highlight key limitations, such as the challenges of generalizing to unseen scenarios and relying on static \gls{lc} data. Addressing these issues will be essential for scaling the approach to broader applications.  

% Future Directions  
Future research should focus on integrating dynamic datasets, such as time-varying vegetation indices, climate-related variables, biophysical and deposition products to capture temporal variations in isoprene emissions better. Exploring hybrid architectures that combine the strengths of \gls{hat} and \gls{san} could further enhance model performance across diverse regions. Additionally, expanding the scope of the study to include global datasets and multi-year observations would enable a more comprehensive evaluation of the model's scalability and generalization capabilities.  
\section{Conclusions}
\label{sec:conclusions}
\glsreset{tc}
\glsreset{cl}
\glsreset{lc}
\glsreset{lai}

This study explored the integration of \gls{lc} priors into \gls{dl}-based \gls{sr} frameworks to enhance the spatial resolution of isoprene emission maps.
%over Europe. 
By incorporating \gls{cl} and \gls{tc} data as emission drivers, we demonstrated significant improvements in the accuracy of super-resolved isoprene maps. Notably, we utilized the most up-to-date and high-resolution isoprene inventory over Europe, further enhancing the precision and validity of our results. The findings highlight the critical role of vegetation types in influencing \gls{bvoc} emissions and underscore the potential of \gls{misr} techniques for addressing spatial resolution challenges in atmospheric research.

The 3-channel experiment consistently outperformed other configurations. Including additional semantic priors constrained the ill-posed nature of the \gls{sr} problem, highlighting the value of integrating diverse data sources to guide \gls{sr} processes.

The comparison between the \gls{hat} and \gls{san} models revealed distinct strengths in capturing global versus localized emission patterns, emphasizing the importance of architecture selection based on data characteristics. However, the generalization studies underscored key challenges, including the difficulty of extrapolating to unseen climate zones and geographic regions and the limitations of relying on static \gls{lc} data.

Despite these challenges, the proposed approach provides a valuable foundation for improving \gls{bvoc} emission inventories, with implications for air quality monitoring, climate modeling, and ecological research. Future work should focus on integrating dynamic and diverse datasets, developing hybrid architectures, and expanding the geographical and temporal scope of the study. These efforts will be crucial for advancing the scalability and applicability of \gls{sr} methods in atmospheric and climate sciences.

% {
% \setglossarystyle{altlist} % Single-column, better for narrow terms
% \renewcommand{\glsnamefont}[1]{\mbox{\footnotesize\bfseries #1}}
% \footnotesize
% \printglossary[type=\acronymtype]
% }

\bibliographystyle{IEEEtran}
\bibliography{refs_clean.bib}


%\vfill


\end{document}